\documentclass[10pt,twocolumn,letterpaper]{article}

\usepackage[pagenumbers]{cvpr} 

\usepackage[dvipsnames]{xcolor}

\definecolor{cvprblue}{rgb}{0.21,0.49,0.74}
\usepackage[pagebackref,breaklinks,colorlinks,citecolor=cvprblue]{hyperref}
\usepackage{tikz}
\usepackage{float}
\usepackage{listings}

\def\confName{CVPR}
\def\confYear{2024}

\title{Efficient Quantization Strategies for Latent Diffusion Models}

\author{
Yuewei Yang
\and
Xiaoliang Dai
\and
Jialiang Wang
\and
Peizhao Zhang
\and
Hongbo Zhang
\and
Meta GenAI\\
{\tt\small \{yueweiyang, xiaoliangdai, jialiangw, stzpz, hbzhang\}@meta.com}
}

\begin{document}
\maketitle
\begin{abstract}
Latent Diffusion Models (LDMs) capture the dynamic evolution of latent variables over time, blending patterns and multimodality in a generative system. Despite the proficiency of LDM in various applications, such as text-to-image generation, facilitated by robust text encoders and a variational autoencoder, the critical need to deploy large generative models on edge devices compels a search for more compact yet effective alternatives. Post Training Quantization (PTQ), a method to compress the operational size of deep learning models, encounters challenges when applied to LDM due to temporal and structural complexities. This study proposes a quantization strategy that efficiently quantize LDMs, leveraging Signal-to-Quantization-Noise Ratio (SQNR) as a pivotal metric for evaluation. By treating the quantization discrepancy as relative noise and identifying sensitive part(s) of a model, we propose an efficient quantization approach encompassing both global and local strategies. The global quantization process mitigates relative quantization noise by initiating higher-precision quantization on sensitive blocks, while local treatments address specific challenges in quantization-sensitive and time-sensitive modules. The outcomes of our experiments reveal that the implementation of both global and local treatments yields a highly efficient and effective Post Training Quantization (PTQ) of LDMs.
\end{abstract}
\section{Introduction}
\label{sec:intro}

\begin{figure}[htbp]
    \centering
    \includegraphics[scale=0.23]{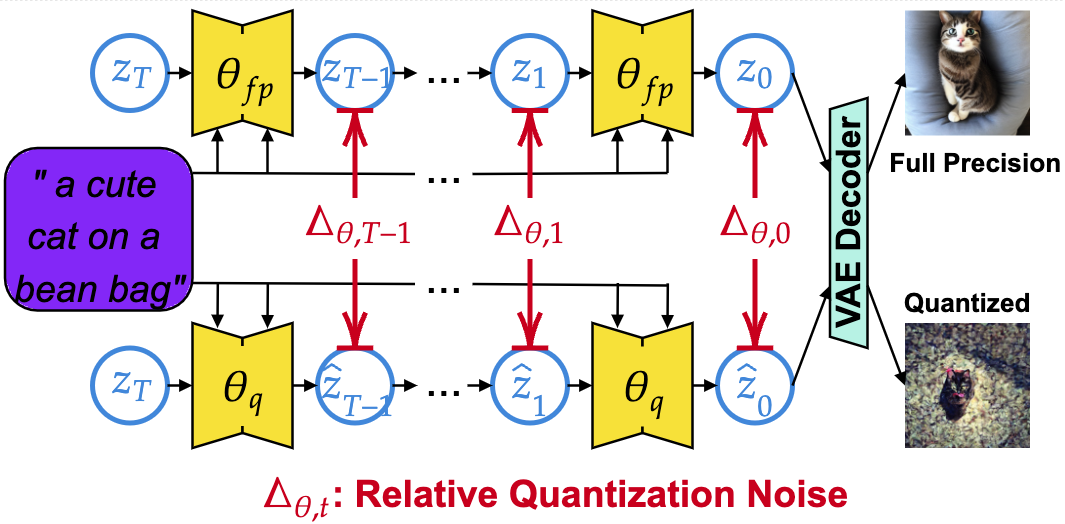}
    \caption{Quantization introduces error between the outputs of full-precision model ($\theta_{fp}$) and quantized model ($\theta_{q}$). Evaluating relative quantization noise at different positions and times ($\Delta_{\theta,t}$) of the reverse diffusion process can identify quantization-sensitive blocks/modules. A more efficient quantization strategy can target identified problems.}
    \label{fig:intro}
\end{figure}

Diffusion models (DMs) have a compelling ability to generate complex patterns via emulating dynamic temporal evolution in a data representation space. Latent diffusion models (LDMs) captures the temporal dynamics in a latent data representation space. With a combination of text encoder(s) and an image variational autoencoder\cite{rombach2022high,podell2023sdxl}, the LDMs outperform other visual generative models in various computer vision applications \cite{kim2022diffusionclip,zhang2023adding,li2023bbdm,tumanyan2023plug,lim2023image}. Within a UNet denoising model \cite{ronneberger2015u}, downsampling blocks reduce the spatial dimensions of the input data to capture high-level features and minimize computational complexity, and upsampling blocks increase the spatial dimensions of the data to recover finer details for more accurate approximation. These blocks in LDMs capture finner and sparse details and generate high-quality outputs. Nonetheless, the generalization of LDMs is hindered by their extensive parameter count, posing challenges for deploying these large generative models on edge devices\cite{zhang2022shifting}. The pressing demand for practical usage on such devices necessitates a more efficient deployment for LDMs.

Among various technique to enable an efficient deployment of large generative models, quantization reduces numerical precision to enhance efficiency, while post-training quantization (PTQ) fine-tunes this process after training for a balanced model size and computational efficiency. PTQ distinguishes itself from other efficiency methods like pruning and distillation \cite{fang2023structural, song2023consistency, salimans2022progressive, meng2023distillation} by avoiding structural modifications to UNet \cite{fang2023structural,li2023efficient,bolya2023token,moon2023early}, bypassing the need for original training data \cite{salimans2022progressive,meng2023distillation}, and eliminating the requirement for model retraining \cite{nichol2021improved,song2020denoising,lu2022dpm}. While quantization on Large Language Models (LLMs) exhibits promising compression results \cite{xiao2023smoothquant,bai2022towards}, its success in language generative models does not seamlessly extend to LDMs due to two main challenges. First, local down-and-up spatial compression are prone to magnify quantization error. Second, global temporally dynamic activation outliers are disposed to impede quantization performance. Prior works \cite{li2023q, he2023ptqd, wang2023towards} address global dynamic activation by selecting uniformly distributed calibration data across inference timesteps, dealing dynamic activation ranges with optimized quantization methods. A more in-depth analysis is needed for an efficient LDM quantization at local and global levels.

In this work, we analyze LDM quantization under the lens of relative quantization noise. We propose an efficient quantization strategy that adeptly identifies blocks or modules sensitive to quantization, enabling the adaptation of tailored solutions for swift and efficient quantization. See Figure \ref{fig:intro} for an overview. The contributions of this work are as following:

\begin{itemize}
    \item This is the first work to propose an efficient quantization strategy that determines effective quantization solutions for LDMs at both global (block) and local (module) levels via analyzing relative quantization noise.
    \item We adapt an efficient metric $\mathbf{SQNR}$ to account for both accumulated global quantization noise and relative local quantization noise.
    \item Globally, we suggest a hybrid quantization approach that adeptly selects a specific LDM block for the initiation of higher-precision quantization, mitigating elevated relative quantization noise.
    \item Locally, our proposal involves the implementation of a smoothing mechanism to alleviate activation quantization noise, achieved through qualitative identification of the most sensitive modules.
    \item We suggest a single-sampling-step calibration, capitalizing on the robustness of local modules to quantization when diffusion noise peaks in the final step of the forward diffusion process. A single-sampling-step calibration remarkably improves the efficiency and performance of quantization.
\end{itemize}

\section{Preliminaries}
\label{sec:preliminaries}
In this section, we expound upon the foundational understanding of LDMs and the concept of Quantization.
\subsection{Latent Diffusion Models}
\label{subsec:LDMs}
DMs \cite{sohl2015deep,ho2020denoising} involve two processes: forward diffusion and reverse diffusion. Forward diffusion process adds noise to input image, $\mathbf{x}_0$ to $\mathbf{x}_T$ iteratively and reverse diffusion process denoises the corrupted data, $\mathbf{x}_T$ to $\mathbf{x}_0$ iteratively. LDMs involve the same two processes but in the latent space $\mathbf{z}_i$ instead of original data space. \textbf{We reiterate the diffusion processes in the latent space}.

The forward process adds a Gaussian noise, $\mathcal{N}(\mathbf{z}_{t-1};\sqrt{1-\beta_t}\mathbf{z}_{t-1},\beta_t \mathbb{I})$, to the example at the previous time step, $\mathbf{z}_{t-1}$. The noise-adding process is controlled by a noise scheduler, $\beta_t$.

The reverse process aims to learn a model to align the data distributions between denoised examples and uncorrupted examples at time step $t-1$ with the knowledge of corrupted examples at time step $t$.

To simplify the optimization, \cite{ho2020denoising} proposes only approximate the mean noise, $\theta(\mathbf{z}_{t},t)\sim \mathcal{N}(\mathbf{z}_{t-1};\sqrt{1-\beta_t}\mathbf{z}_{t},\beta_t \mathbb{I})$, to be denoised at time step $t$ by assuming that the variance is fixed. So the reverse process or the inference process is modeled as:
\begin{align}
    \begin{split}
        unconditional:p_\theta(\mathbf{z}_{t-1} \vert \mathbf{z}_t) &= \theta(\mathbf{z}_{t},t) \\
        conditional:p_\theta(\mathbf{z}_{t-1} \vert \mathbf{z}_t) &= \theta(\mathbf{z}_{t},t,\tau(y))
    \end{split}
\end{align}

\begin{figure}[htbp]
    \centering
    \begin{subfigure}[b]{0.2\textwidth}
         \centering
         \includegraphics[width=\textwidth]{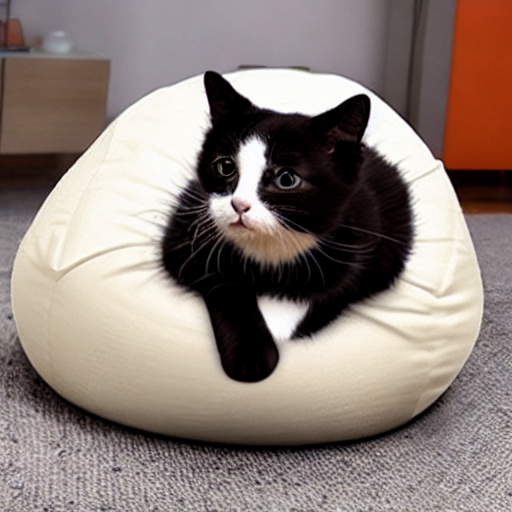}
         \caption{Full Precision}
         \label{fig:exm_fp}
     \end{subfigure}
     \hfill
     \begin{subfigure}[b]{0.2\textwidth}
         \centering
         \includegraphics[width=\textwidth]{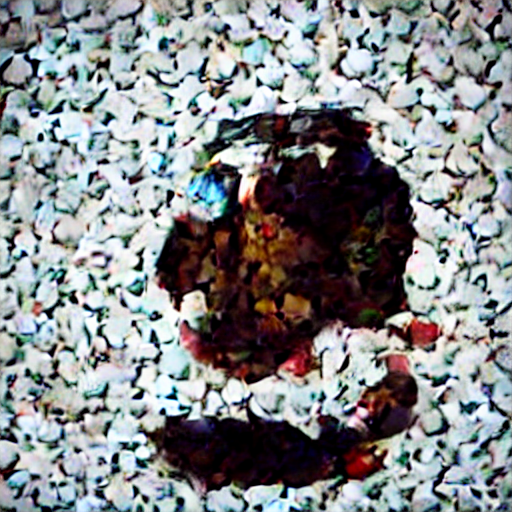}
         \caption{Min-Max Quantized}
         \label{fig:exp_mmq}
     \end{subfigure}
     \caption{Examples of full-precision and min-max quantization.}
     \label{fig:examples}
\end{figure}

\subsection{Quantization}
\label{subsec:prelim_quantization}
Post-training quantization (PTQ) reduces numerical representations by rounding elements $v$ to a discrete set of values, where the quantization and de-quantization can be formulated as:
\begin{equation} \label{eq:quantdequant}
    \hat{v} = s\cdot{clip}(round(v/s),c_{min},c_{max})
\end{equation}
where $s$ denotes the quantization scale parameters. round(·) represents a rounding function \cite{cai2020zeroq,wu2020easyquant}. $c_{min}$ and $c_{max}$ are the lower and upper bounds for the clipping function $clip(\cdot)$. Calibrating parameters in the PTQ process using weight and activation distribution estimation is crucial. A calibration relying on min-max values is the most simple and efficient method but it leads to significant quantization loss (see examples in Figure \ref{fig:examples}), especially when outlier values constitute a minimal portion of the range \cite{wei2022outlier}. More complicated optimized quantization is applied to eliminate the outliers for better calibration. For example, PTQ4DM \cite{shang2023post}
and Q-Diffusion \cite{li2023q} apply reconstruction-based PTQ methods \cite{li2021brecq} to diffusion models. PTQD \cite{he2023ptqd} further decomposes quantization noise and fuses it with diffusion noise. Nevertheless, all these methods demand a significant increase in computational complexity. In this study, our exclusive experimentation revolves around min-max quantization due to its simplicity and efficiency. Additionally, we aim to demonstrate how our proposed strategy proficiently mitigates the quantization noise inherent in min-max quantization.

\section{Method}
\label{sec:method}

\subsection{Quantization Strategy}
\label{subsec:strategy}
\begin{figure}[htbp]
     \centering
     \begin{subfigure}[b]{0.45\textwidth}
         \centering
         \includegraphics[width=\textwidth]{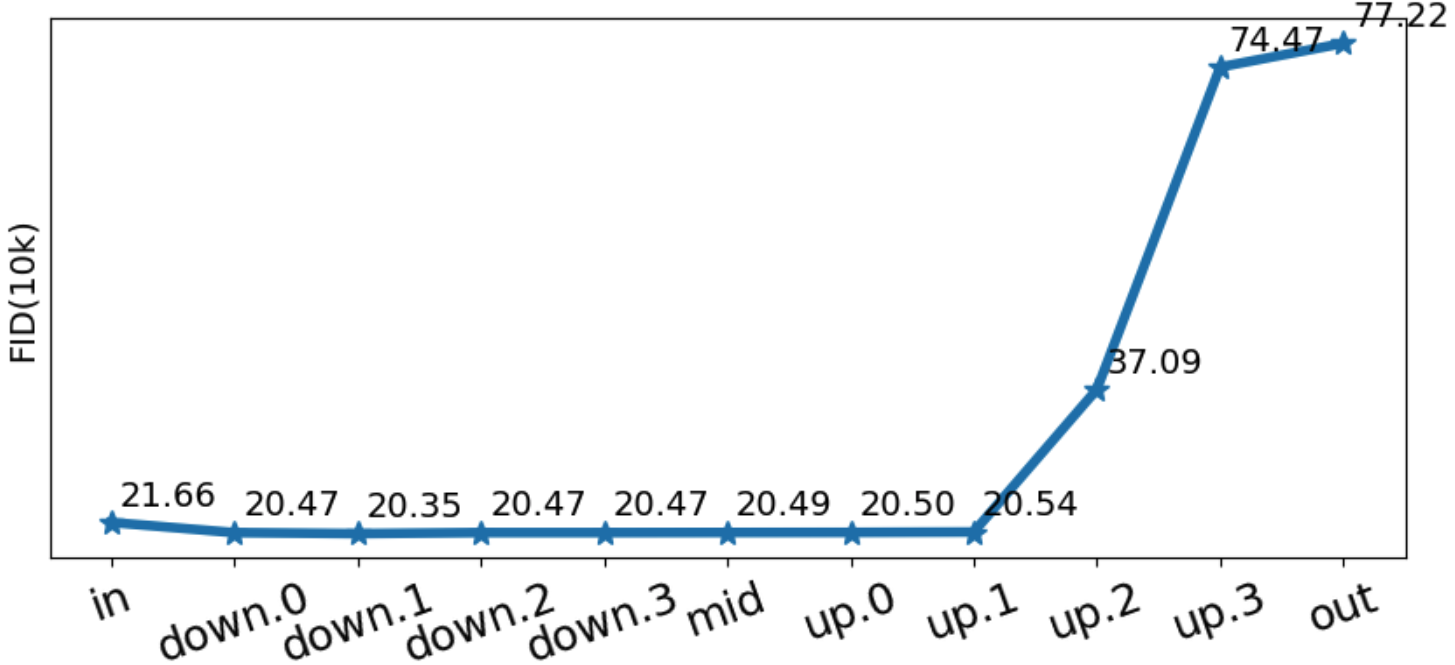}
         \caption{Progressively quantizing end blocks corrupts image quality.}
         \label{fig:global fids}
     \end{subfigure}
     \hfill
     \begin{subfigure}[b]{0.45\textwidth}
         \centering
         \includegraphics[width=\textwidth]{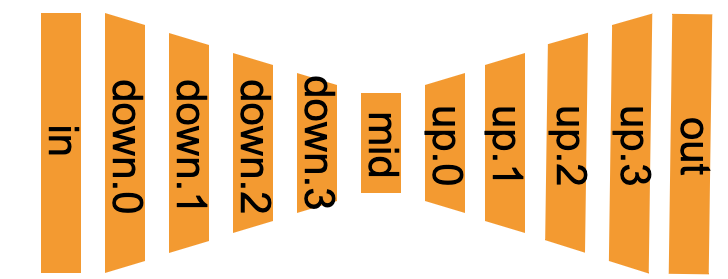}
         \caption{Blocks in UNet LDM 1.5.}
         \label{fig:blocks}
     \end{subfigure}

     \caption{We observe that end blocks are sensitive to homogeneous quantization. A hybrid quantization strategy should target these blocks efficiently.}
     \label{fig:strategy}
\end{figure}

In general, a homogeneous quantization strategy is applied to all parts in a network for its simplicity. In other words, the same precision quantization is applied to all modules. Though different quantization schemes coincide in mixed-precision quantization \cite{chen2021towards,dong2019hawq,wang2019haq} and non-uniform quantization \cite{li2019additive,zhang2018lq,jeon2022mr}, additional training or optimization is required to configure more challenging quantization parameters. LDMs consist of many downsampling and upsampling blocks (Figure \ref{fig:blocks}). The concatenated blocks in LDMs pose challenges for end-to-end quantization, given dynamic activation ranges and recursive processing across steps and blocks. In contrast to prior attempts to uniformly quantize all blocks, our approach suggests an efficient quantization strategy by initiating a hybrid quantization at a specific block and subsequent blocks.

We attentively discover that quantizing up to a specific block in LDM imposes minimum loss in image generating quality. In Figure \ref{fig:global fids}, as we measure FID for a progressive homogeneous quantization on LDM 1.5, blocks nearer to the output exhibit higher sensitivity to quantization. Identifying these sensitive blocks beforehand at a low cost facilitates the implementation of a decisive strategy. However, computing FID is computationally expensive and time-consuming. An efficient metric is required to pinpoint sensitive blocks (and modules) for the development of an improved quantization strategy.

\subsection{Relative Quantization Noise}
\label{subsec:relative}

We denote $\theta_{fp}$ and $\theta_{q}$ are the full-precision model and quantized model respectively. If we simplify the iterative denoising process as (\ref{eq:simple}), where the timestep embedding and text embedding are omitted for simplicity, intuitively each parameterized module contributes quantization noise and the recursive process will only accumulate these noise. This behaviour is modeled as (\ref{eq:error}) where $\Phi$ is an accumulative function. This is akin to conducting sensitivity analysis to scrutinize how the accumulation of quantization impact on each block or module and influence the quality of the generated images.

\begin{equation} \label{eq:simple}
   \mathbf{z}_0=\Pi_{t=T}^{0}\theta_{fp}(\mathbf{z}_T), \mathbf{\hat{z}}_0=\Pi_{t=T}^0\theta_{q}(\mathbf{z}_T)
\end{equation}
\begin{equation} \label{eq:error}
\begin{split}
    \Delta_{\theta,T-1} &= metric(\theta_{fp}(\mathbf{z}_T),\theta_{q}(\mathbf{z}_T)) \\
    \Delta_{\theta,t-1} &= \Phi_{t=T-1}^{t-1}(\Delta_{\theta,t}) \\
\end{split}
\end{equation}

We need a computationally efficient $metric(.,.)$ function to identify sensitive blocks in $\theta$ if the accumulated $\Delta_{block,t:T\rightarrow1}$ is too high, and identify sensitive modules if the relative $\Delta_{module}$ is too high. In other words, the $metric(.,.)$ necessitates three key properties:
\begin{itemize}
    \item \textbf{Accumulation} of iterative quantization noise needs to be ensured as the global evaluation of the quantization noise at the output directly reflect image quality.
    \item \textbf{Relative} values across different modules provide qualitative comparisons to identify local sensitivity.
    \item \textbf{Convenient} computation allows an efficient evaluation and analysis which expedites quantization process.
\end{itemize}
Hence metrics like MSE is not sufficient to make realtive and fair local comparisons, and FID is not efficient to facilitate the quantization process.

\begin{figure}[htbp]
    \centering
    \includegraphics[scale=0.18]{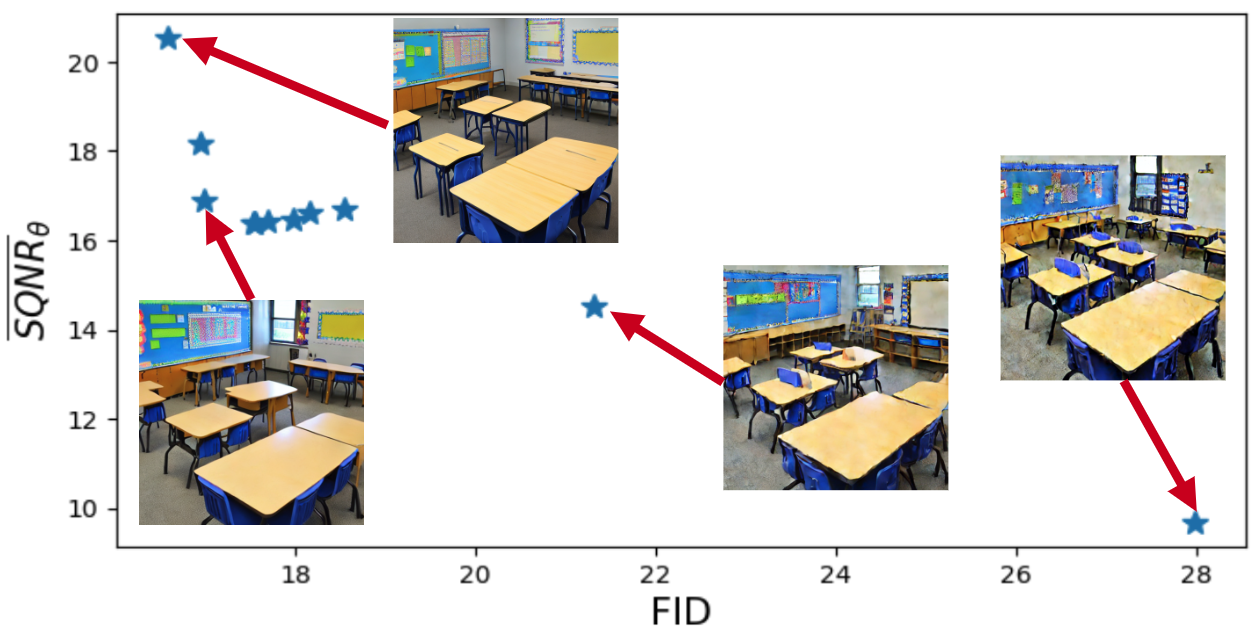}
    \caption{A strong correlation between $\mathbf{SQNR}$ and FID. High quantization noise results in lossy images.}
    \label{fig:sqnr_fid}
\end{figure}

We adapt a relative distance metric, $\mathbf{SQNR}$ \cite{lathi1990modern}, as the $metric(.,.)$:
\begin{equation}
    \mathbf{SQNR}_{\xi,t} = 10log\mathbb{E}_{\mathbf{z}}\frac{||\xi_{fp}(\mathbf{z}_t)||^2_2}{||\xi_{q}(\mathbf{\hat{z}}_t)-\xi_{fp}(\mathbf{z}_t)||^2_2}
\end{equation}
where $\xi$ represents the whole model for global sensitivity evaluation and a module for local sensitivity evaluation.
Other works have associated $\mathbf{SQNR}$ with the quantization loss. \cite{sheng2018quantization} evaluates $\mathbf{SQNR}$ for quantization loss on a convolutional neural network (CNN). \cite{lee2018quantization} optimizes $\mathbf{SQNR}$ as the metric to efficiently learn the quantization parameters of the CNN. \cite{pandey2023softmax} computes $\mathbf{SQNR}$ at the end of inference steps to quantify the accumulated quantization loss on softmax operations in diffusion models. \cite{pandey2023practical} assesses $\mathbf{SQNR}$ to fairly compare the bit options for the quantized weight in a deep neural network. Therefore $\mathbf{SQNR}$ is an optimal candidate for $metric(.,.)$ and also it offers three major advantages:

\begin{itemize}
    \item At global levels, instead of modeling an accumulative function $\Phi$, we evaluate the time-averaged $\mathbf{SQNR}$ at the output of the $\theta$, i.e. $\Delta_{block_{N},t:T\rightarrow1}=\mathbb{E}_{t=1}^{T}[\mathbf{SQNR}_{\theta,t}]$. Averaging the relative noise at the output comprises an end-to-end accumulation nature. A low $\Delta_{block_{N},t:T\rightarrow1}$ indicates a high quantization noise, which leads to poorly generated images. We justify this correlation as qualitative measure $\mathbb{E}_{t=1}^{T}[\mathbf{SQNR}_{\hat{\theta},t}]$ and FID and show the correlation in Figure \ref{fig:sqnr_fid}.
    \item At local levels, $\mathbf{SQNR}_{\xi,t}$ is a \textbf{relative} metric that measures the noise ratio between the full-precision and quantized values. This aids a fair comparison between local modules to identify the sensitive modules.
    \item A \textbf{small number} of examples is sufficient to compute the $\mathbf{SQNR}$ instantaneously with high confidence (Supplementary \ref{supp:computesqnr}). This swift computation enables an efficient assessment of quantization noise and the quality of the generated image, in comparison to FID and IS.
\end{itemize}

\subsection{Block Sensitivity Identification and Global Hybrid Quantization}
\label{subsec:global}

\begin{figure}[htbp]
     \centering
     \begin{subfigure}[b]{0.46\textwidth}
         \centering
         \includegraphics[width=\textwidth]{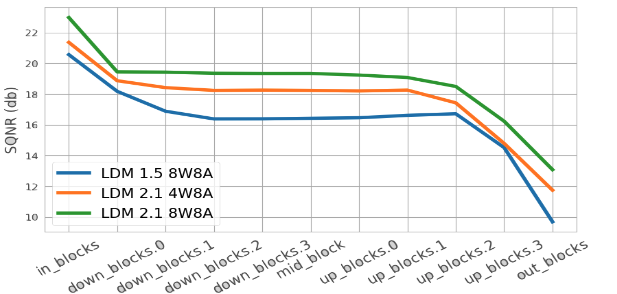}
         \caption{Global Blocks Sensitivity: Sensitivity starts to descend at a upsampling block.}
         \label{fig:block_sensitivity}
     \end{subfigure}
     \hfill
     \begin{subfigure}[b]{0.46\textwidth}
         \centering
         \includegraphics[width=\textwidth]{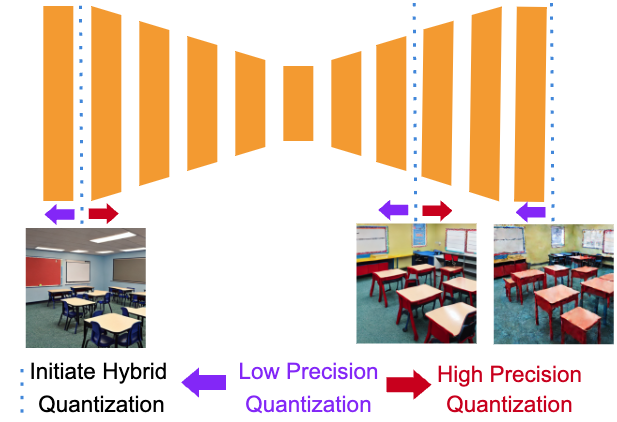}
         \caption{Global Strategy}
         \label{fig:global strategy}
     \end{subfigure}
     \caption{\textbf{Global hybrid quantization}: Block Sensitivity \ref{fig:block_sensitivity} identifies sensitive \textcolor{BrickRed}{blocks} by locating descending $\mathbf{\overline{SQNR}}_{\theta}$. Block sensitivity matches the observation in \ref{subsec:strategy} that end blocks are more sensitive to quantization. Initiate a higher precision quantization on sensitive blocks.}
     \label{fig:global_strategy}
\end{figure}

\begin{equation}\label{eq:avg_sqnr}
    \overline{\mathbf{SQNR}}_{\theta}=\mathbb{E}_{t=1}^{T}[\mathbf{SQNR}_{\theta,t}]
\end{equation}
Conventionally, the quantization is applied with a consistent precision on the whole model. But the quantization fails with low-bit precision constantly \cite{li2023q,he2023ptqd} and there is no substantiate evidence to identify the root cause for these failures. We measure the time-averaged $\mathbf{SQNR}$ at the output of $\theta$ which is denoted as $\mathbf{\overline{SQNR}}_{\theta}$ shown in Eq (\ref{eq:avg_sqnr}). Progressively quantizing each block in $\theta$ examines the sensitivity of each block based on $\mathbf{\overline{SQNR}}_{\theta}$.  Once blocks with high sensitivity are identified, a higher-precision quantization can be applied to alleviate quantization noise from these blocks. The proposed global hybrid quantization is illustrated in Figure \ref{fig:global strategy}. We first explore the sensitivity of each block in the UNet by quantizing first $N$ blocks, and leave the rest blocks unquantized. In Figure \ref{fig:block_sensitivity}, the block sensitivity of different LDMs at different quantization bits consistently indicates that latter blocks in up blocks are severely sensitive to quantization. Using the time-averaged $\mathbf{SQNR}$, sensitive blocks can be identified efficiently and a higher precision quantization can be applied on these blocks collectively.

\subsection{Module Sensitivity Identification and Local Noise Correction}
\label{subsec:local}

\begin{figure}[htbp]
     \centering
    \includegraphics[scale=0.18]{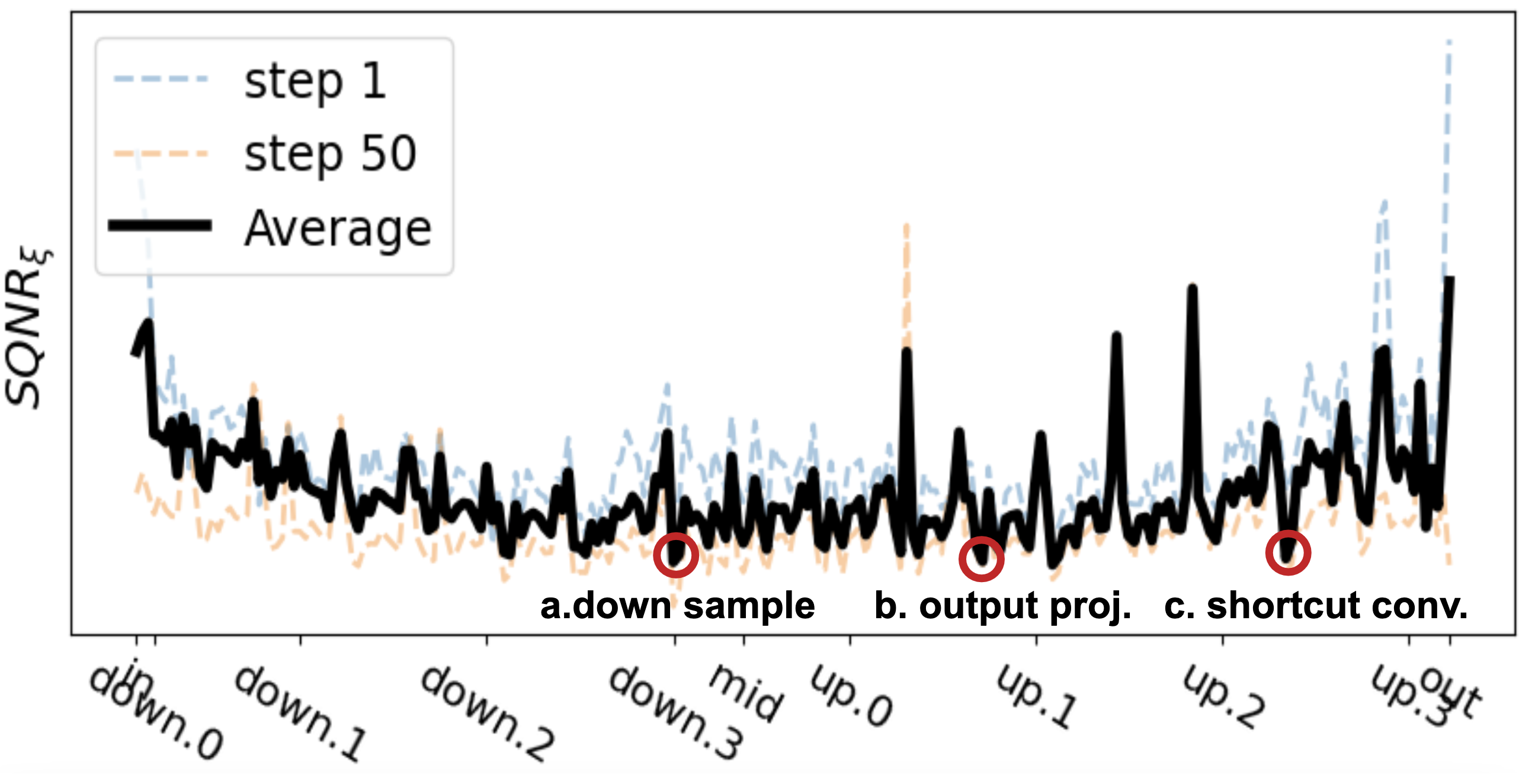}
     \caption{\textbf{Local Sensitivity Identification}: Three \textcolor{BrickRed}{red} circles identify three sensitive modules with low $\mathbf{SQNR}_{\xi}$. Low $\mathbf{SQNR}_{xi}$ indicates more sensitive to quantization. $\mathbf{SQNR}_{\xi}$ for steps$\in[1,50]$ are show.}
     \label{fig:local_noise}
\end{figure}

Similarly to block sensitivity analysis, we evaluate $\mathbf{SQNR}_{\xi}$ at each module and visualize the sensitivity at different inference steps. We intend to formulate a local sensitivity first and identify local sensitive modules as illustrated in Figure \ref{fig:local_noise}. We visualize the $\mathbf{SQNR}_{\xi}$ for each module at two inference steps, $[1,50]$, and the average over all 50 sampling steps when quantizing LDM 1.5 with 8A8W in Figure \ref{fig:local_noise} (Illustration for other LDMs in Supplementary \ref{supp:modulesensitivity}). 
A low local $\mathbf{SQNR}_{xi}$ only suggests that one module is sensitive to quantization noise, and vice versa. From the module sensitivity across inference time and across the whole model, two types of root causes for the local sensitivity can be identified and will be addressed respectively.

\subsubsection{Quantization Corrections to Sensitive Modules}
\begin{figure}
     \centering
     \begin{subfigure}[b]{0.44\textwidth}
         \centering
         \includegraphics[width=0.8\textwidth]{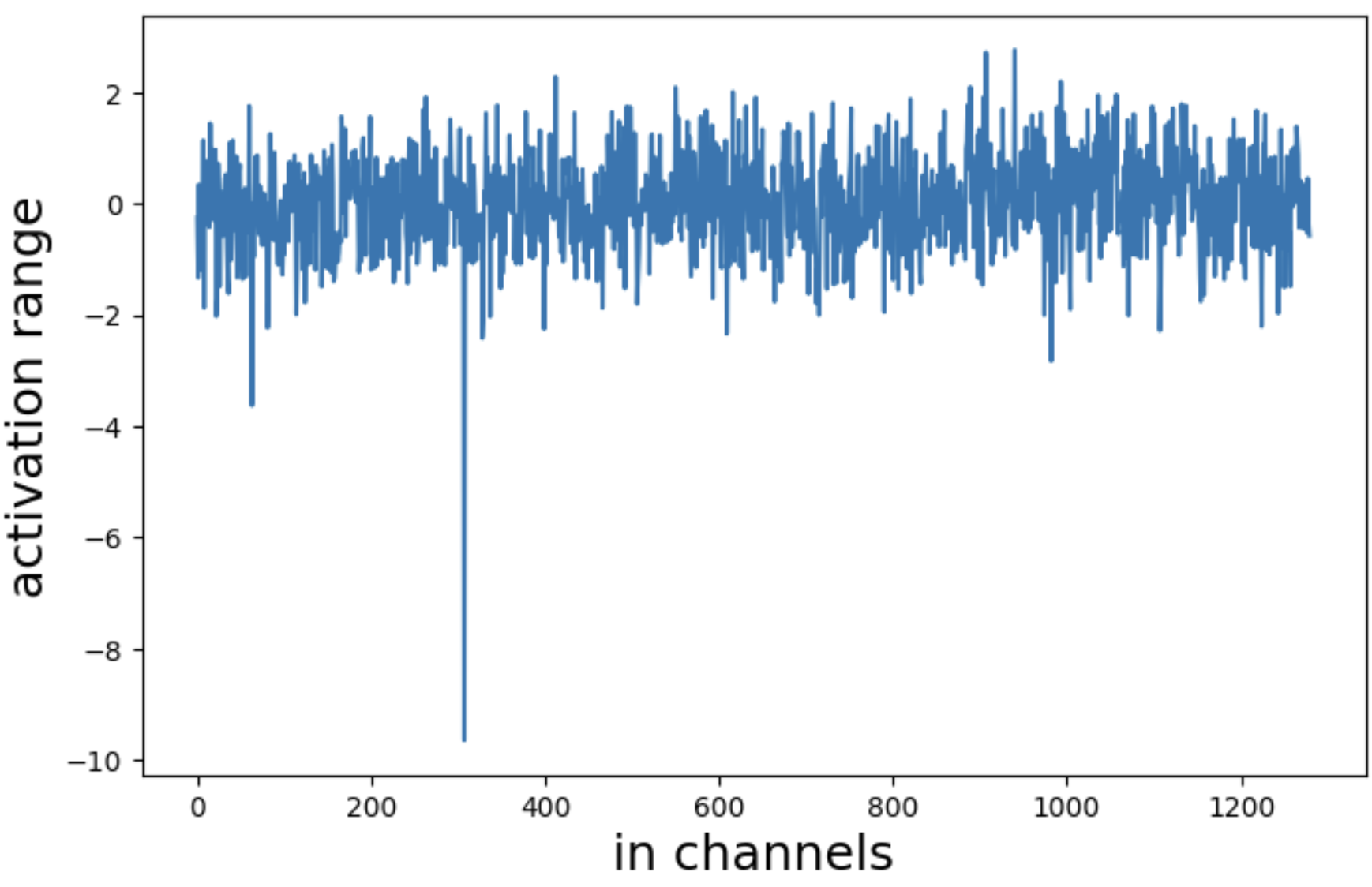}
         \caption{Spatial Downsampling.}
         \label{fig:act1}
     \end{subfigure}
     \hfill
     \begin{subfigure}[b]{0.44\textwidth}
         \centering
         \includegraphics[width=0.8\textwidth]{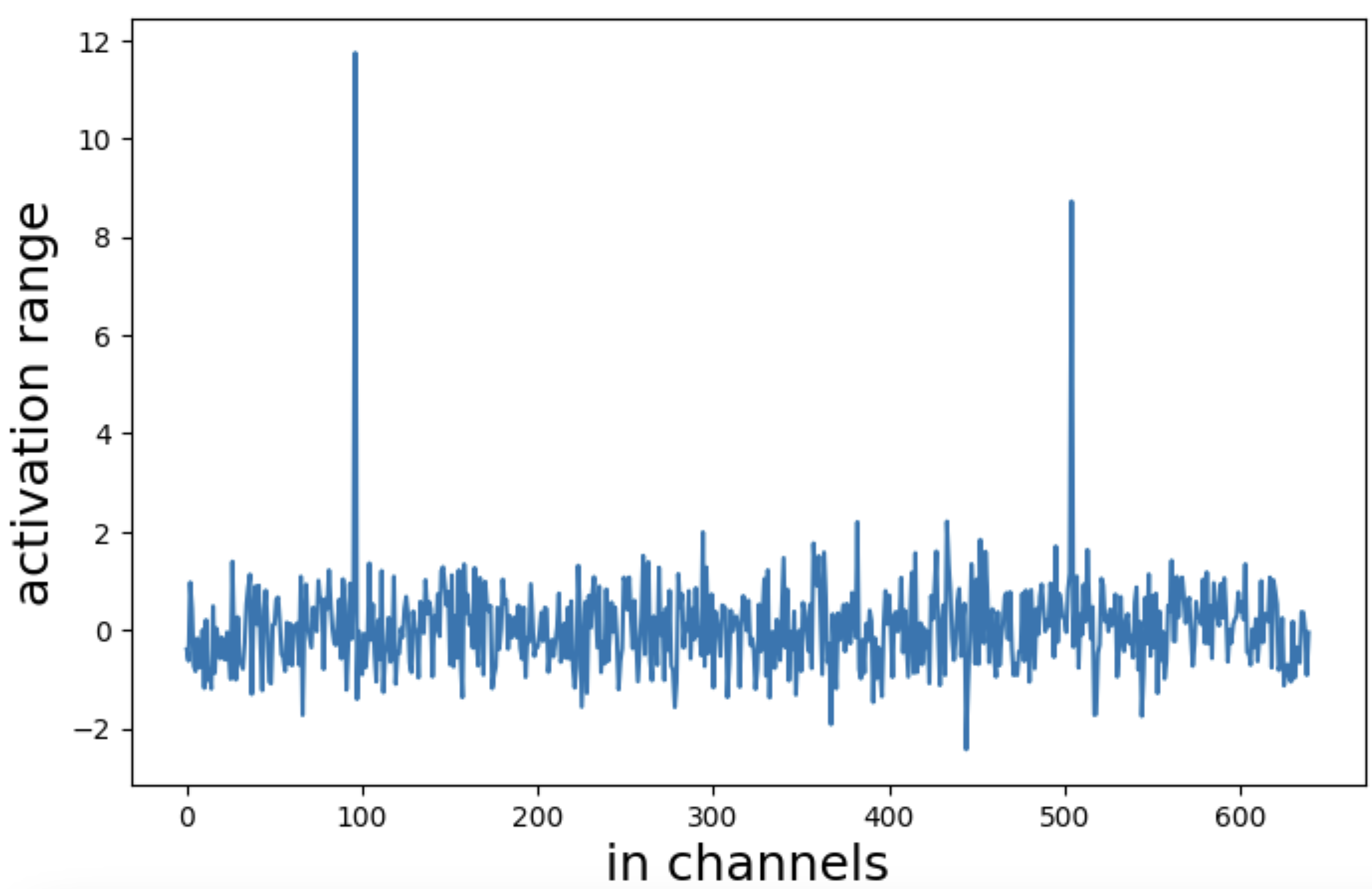}
         \caption{Transformer Ouput Projection.}
         \label{fig:act2}
     \end{subfigure}
     \begin{subfigure}[b]{0.44\textwidth}
         \centering
         \includegraphics[width=0.8\textwidth]{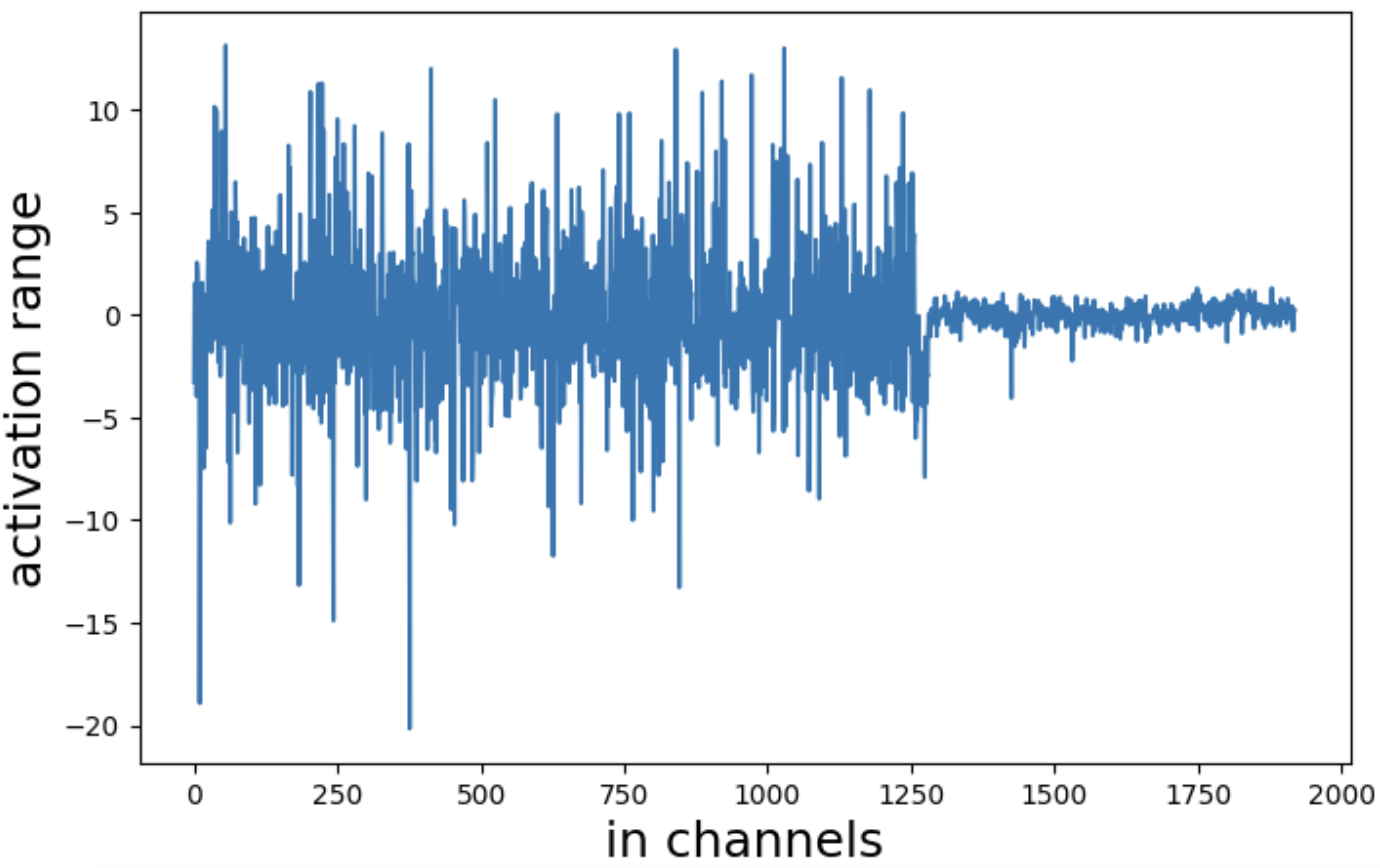}
         \caption{Shortcut Connection.}
         \label{fig:act3}
     \end{subfigure}
     \hfill
     \caption{Activation ranges at three identified most sensitive modules as red circled in Figure \ref{fig:local_noise}. Outliers in certain channels make the quantization difficult.See Supplementary \ref{supp:outliers} for 3d visualization.}
     \label{fig:act}
\end{figure}

The first root cause for the high quantization sensitivity can be identified by listing modules with the lowest $\mathbf{SQNR}_{\xi}$. In fact, three types of operations are consistently sensitive to quantization: a).Spatial sampling operations; b). Projection layers after the attention transformer block; c).Shortcut connection in up blocks. (See Supplementary \ref{supp:modulesensitivity} for details).
Though c). is addressed in \cite{li2023q} via splitting connections, we analyse all sensitive modules with relative noise and provide a uniform solution. After identifying these modules, the activation ranges can be collected across inference steps. In Figure \ref{fig:act} we visualize the average activation ranges over output channels operations a) and b). It is obvious that there are outliers in some channels in the activation. As discussed in Section \ref{sec:intro}, these outliers pose difficult challenges to quantization as they stretch the activation range. An apparent solution is to use per-channel activation quantization on these sensitive operations, but per-channel activation cannot be realized with any hardware \cite{xiao2023smoothquant,krishnamoorthi2018quantizing,banner2019post}. We adapt SmoothQuant \cite{xiao2023smoothquant} to resolve the outlier quantization challenges for selected sensitive modules:

\begin{equation} \label{eq:smoothquant}
    \begin{split}
        \mathbf{\hat{Z}}\mathbf{\hat{W}} &= (\mathbf{Z}diag(\mathbf{s})^{-1})(diag(\mathbf{s})\mathbf{W})\\
        \mathbf{s}_j &= max|\mathbf{Z}_j|^{\alpha}/max|\mathbf{W}_j|^{1-\alpha}
    \end{split}
\end{equation}
where $\mathbf{Z}\in{\mathcal{R}^{C\times{(H,H\times{W})}}}$ is the input activation to a parameterized module with weight $\mathbf{W}\in{\mathcal{R}^{C\times{C'}(\times{n}\times{n})}}$. $\mathbf{s}_j,j\in{[1,C]}$ is the scale to mitigate the outliers in channel $j$ and migrate the quantization burden to weight with a migration factor $\alpha$. Though $diag(\mathbf{s})$ can be fused to the parameterized module weights for efficient computation, $diag(\mathbf{s}^{-1})$ cannot be fused to previous layers as proposed in \cite{xiao2023smoothquant}, which means there is a trade-off between the efficiency and performance. We examine this trade-off in section \ref{sec:ablations} via computing the number of total operations and the proportion of sensitive modules to apply SmoothQuant.

\subsubsection{Quantization Sensitivity to Diffusion Noise}
\label{subsub:steps}
\begin{figure}
    \centering
    \begin{subfigure}[b]{0.2\textwidth}
         \centering
         \includegraphics[width=\textwidth]{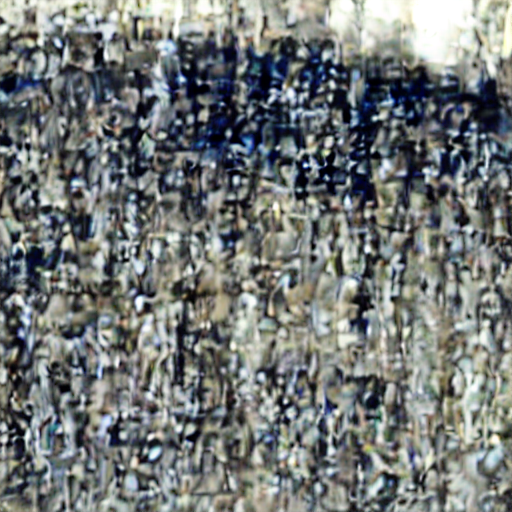}
         \caption{50 steps.}
         \label{fig:50steps}
     \end{subfigure}
     \hfill
     \begin{subfigure}[b]{0.2\textwidth}
         \centering
         \includegraphics[width=\textwidth]{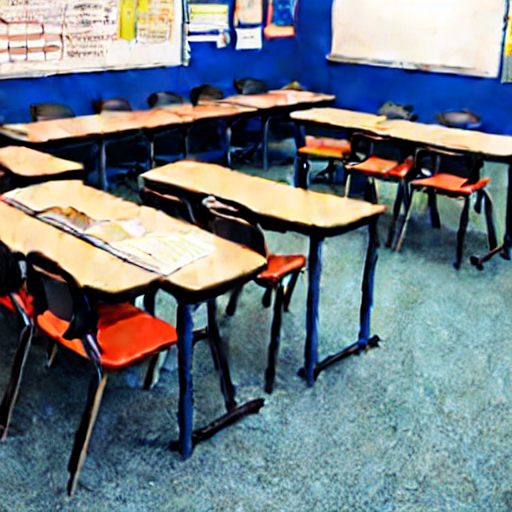}
         \caption{1 step.}
         \label{fig:1step}
     \end{subfigure}
    \caption{Difference between calibrating through 50 steps and 1 step.}
    \label{fig:calibrate}
\end{figure}

The second root cause for the high quantization sensitivity can be identified as the $\mathbf{SQNR}$ difference between the first inference step and the last inference step (dash lines in Figure \ref{fig:local_noise}). For all modules, this discrepancy suggests that the quantization parameters are most robust when the diffusion noise is scheduled at maximum. In general, the quantization parameters are calibrated via inferencing a number of steps with any noise schedulers. As demonstrated in section \ref{sec:preliminaries}, the noise scheduler induces gradually changing noise at each forward diffusion step. Hence the activation range is more dynamic when calibrating across multiple steps. This coincides with the problems identified in \cite{li2023q,he2023ptqd,shang2023post}. This variation causes high quantization noise. Unlike the proposed solutions in prior works, we proposed to only calibrate with a single step, as the variation in maximized diffusion noise will be constrained so that a tighter $c_{min},c_{max}$ can be calibrated for better and faster quantization. See Figure \ref{fig:calibrate} for examples.

\section{Experiment Setup and Evaluation}
\label{sec:expsetup}
\noindent\textbf{Dataset and Quantization Setting} We evaluation our proposed quantization procedures in conditional text-to-image generation. Unconditional calibration over 1k images is performed (conditional calibration aggravates the quantization loss as difference between conditional and unconditional features induces further quantization noise). The evaluation dataset is random 10k generations from MS-COCO validation \cite{lin2014microsoft} as conducted in \cite{li2023q,gu2022vector,cao2018hashgan}. The classifier-free guidance scale is set to 7.5. We only quantize parameterized modules (convolutional layers and fully connected layers). For all quantization, we apply only min-max symmetric quantization scheme to highlight the raw improvement of proposed procedures.
Two quantization settings will be explored 8W8A, 4W8A for low-bit and lower-bit quantization demonstrations. In the global hybrid quantization, fp16 quantization will be applied to sensitive blocks. The default configurations for LDMs are listed in Supplementary

\noindent\textbf{Evaluation Metrics} For each experiment, we report the FID \cite{heusel2017gans} and $\mathbf{\overline{SQNR}}_{\theta}$ to evaluate the performance on generating images of size 512 by 512 for LDM 1.5, LDM 2.1, 1024 by 1024 for LDM XL. All results on baselines and proposed procedures are controlled by a fixed random seed to ensure reproducibility of results. To evaluate the computational efficiency, we calculate Bit Operations (BOPs) for a conditional inference step through the diffusion model using the equation $BOPs = MACs\times{bw}\times{ba}$, where $MACs$ denotes Multiply-And-Accumulate operations, and $bw$ and $ba$ represent the bit width of weights and activations, respectively, as stated in \cite{he2023ptqd,wang2020differentiable}.

\noindent\textbf{Statistics Preparation} It is optional to collect relative quantization noise after quantizng a LDM model as the analysis is consistent across different models and quantization settings as suggested in section \ref{sec:method}. However, for a general LDM it is advised to quantize the model with a specific quantization setting to perform the analysis. For the local outlier correction, the maximum of the absolute activation range needs to be recorded first to calculate $\mathbf{s}_j$ in (\ref{eq:smoothquant}).

\subsection{Results and Discussions}

\subsubsection{Global Hybrid Quantization}
To validate the generalization of proposed quantization strategy, we evaluate the global hybrid quantization procedures on three different LDM models: LDM 1.5, 2.1 base, XL 1.0 base. Note that we only employ single-sampling-step calibration for the hybrid quantization and the smoothing mechanism is not included for better efficiency.

\begin{table}[htbp]
\resizebox{\columnwidth}{!}{
\begin{tabular}{cccccc}
\hline
Method  & Bits(W/A)     & Size(Mb) & TBops & FID$\downarrow$ & $\mathbf{\overline{SQNR}}_{\theta}$(db)$\uparrow$ \\ \hline
        & 32fp/32fp     & 3278.81  & 693.76   & 18.27              & -                 \\
LDM 1.5 & 8/8+16fp/16fp & 906.30       & 99.34    & \textbf{16.58}              & 20.45                 \\
        & 4/8+16fp/16fp & 539.55       & 86.96    & 21.76              & 18.17                \\ \hline
        & 32fp/32fp     & 3303.19  & 694.77    & 19.03              & -                 \\
LDM 2.1 & 8/8+16fp/16fp & 913.80       & 99.45    & \textbf{18.52}              & 19.26                 \\
        & 4/8+16fp/16fp & 544.70       & 87.04    & 21.94              & 18.48                 \\ \hline
        & 32fp/32fp     & 9794.10 & 6123.35    & 19.68               & -                 \\
LDM XL  & 8/8+16fp/16fp & 2562.26       & 604.06    & \textbf{16.48}              &     20.66             \\
        & 4/8+16fp/16fp & 1394.30       & 449.28    & 21.55              & 18.50                 \\ \hline
\end{tabular}}
\caption{Global Hybrid Quantization based on different LDMs.}
\label{tbl:globalhybrid}
\end{table}

\begin{figure*}[htbp]
     \centering
     \begin{subfigure}[b]{0.31\textwidth}
         \centering
         \includegraphics[width=\textwidth]{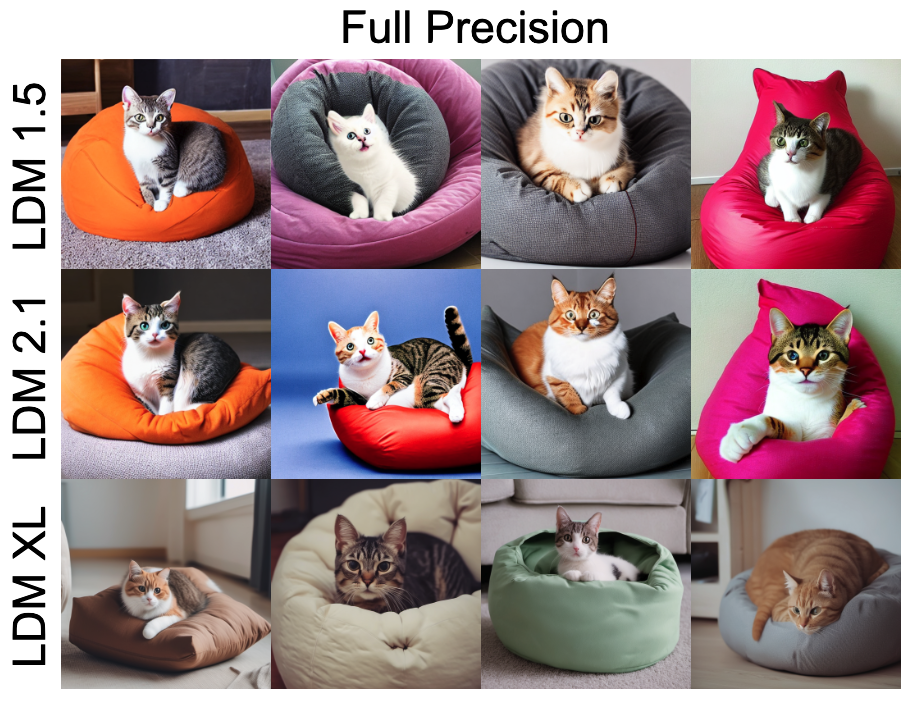}
     \end{subfigure}
     \hfill
     \begin{subfigure}[b]{0.29\textwidth}
         \centering
         \includegraphics[width=\textwidth]{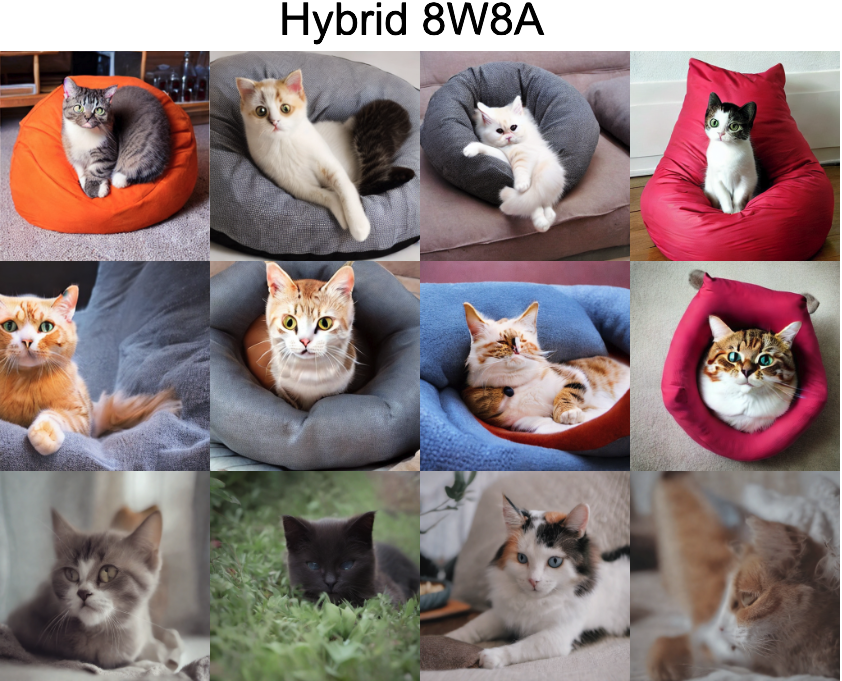}
     \end{subfigure}
     \hfill
     \begin{subfigure}[b]{0.29\textwidth}
         \centering
         \includegraphics[width=\textwidth]{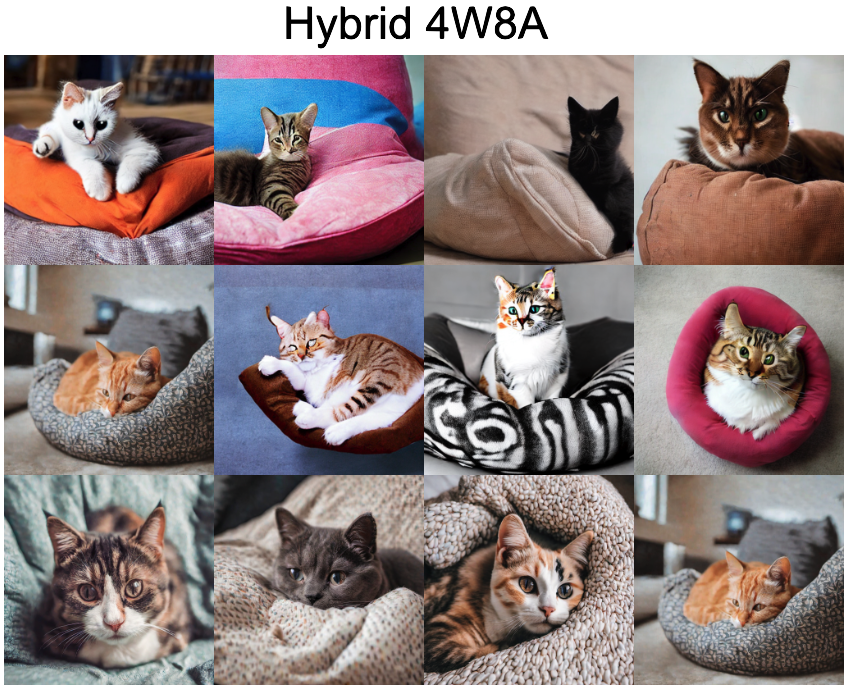}
     \end{subfigure}
        \caption{Global Hybrid Quantization Strategy.}
        \label{fig:global_examples}
\end{figure*}

We report the best configuration on hybrid quantization.
Besides shrinking the operational size of the LDM models, the hybrid quantization significantly improves computational efficiency. For all LDMs, the upsampling blocks that are closer to the output are most sensitive to the quantization. Even though the process is recursive, these blocks contribute most quantization noise at the end of each sampling step. By replacing these blocks with higher-precision quantized blocks, these hybrid blocks filter out most quantization noise which is illustrated as an increase in $\mathbf{\overline{SQNR}}_{\theta}$ shown in the ablation. Interestingly, the FID suggests that the hybrid quantized LDMs is on-par or even better than the full-precision models. The unreliability of FID to quantify quality of image perception is mentioned in other works \cite{chong2020effectively,naeem2020reliable}. But $\mathbf{\overline{SQNR}}_{\theta}$ aligns with the intuition that lower bit quantization accumulates higher quantization noise and hence generates more lossy images.

\subsubsection{Local Noise Correction}
We evaluate the sole power of the proposed local noise corrections on two different LDM models: LDM 1.5, XL 1.0 base.
\begin{table}[htbp]
\resizebox{\columnwidth}{!}{
\begin{tabular}{lccccc}
\hline
\multicolumn{1}{c}{Method} & Bits(W/A)                         & Size(Mb) & TBops   & FID$\downarrow$ & $\mathbf{\overline{SQNR}}_{\theta}$(db)$\uparrow$ \\ \hline
\multicolumn{1}{c}{}       & 32fp/32fp                         & 3278.82  & 693.76  & 18.27              & -                 \\
\multicolumn{1}{c}{}       & 8/8                               & 820.28       & 43.63      & 213.72              & 11.04                 \\
LDM 1.5                    & + corrections                     & 820.28       & 43.63     & 24.51              & 17.78                 \\
                           & 4/8                               & 410.53       & 21.96      & 296.45              & 9.73                 \\
                           & \multicolumn{1}{l}{+ corrections} & 410.53       & 21.96      & 57.71              & 15.71                 \\ \hline
\multicolumn{1}{c}{}       & 32fp/32fp                         & 9794.10 & 6123.35 & 19.68              & -                 \\
                           & 8/8                               & 2450.21       & 385.13      & 76.90              & 16.57                 \\
LDM XL                     & \multicolumn{1}{l}{+ corrections} & 2450.21       & 385.13      & 22.33              & 18.21                 \\
                           & 4/8                               & 1226.23       & 193.85      &    237.40           & 12.29                 \\
                           & \multicolumn{1}{l}{+ corrections} & 1226.23       & 193.85      &     44.41          & 13.71                 \\ \hline
\end{tabular}}
\caption{Local Noise Correction on different LDM models.}
\label{tbl:localcorrection}
\end{table}
To demonstrate the benefit of local quantization strategy, we conduct experiments using the local noise correction on the identified sensitive modules and the single-sampling-step calibration without hybrid global quantization. We adapt SmoothQuant to the $10\%$ most sensitive modules to balance the increased computation and reduced quantization noise. Since we only calibrate quantization parameters with min-max values, the end-to-end quantized baselines generate very noisy images as high FID and low $\mathbf{\overline{SQNR}}_{\theta}$ listed in Table \ref{tbl:localcorrection}. However, with both local quantization strategy implemented, the generated images restore its quality to a large extend. Though we notice there is a gap to the full-precision generated images, the qualitative examples demonstrate much coherent and clear images after the local noise corrections. Note that optimized quantization calibration methods like entropy-based, percentile-based, and BRECQ can further improve the quality. (See Supplementary \ref{supp:quantizeoptimize}.)

\begin{figure*}[h]
     \centering
     \begin{subfigure}[b]{0.235\textwidth}
         \centering
         \includegraphics[width=\textwidth]{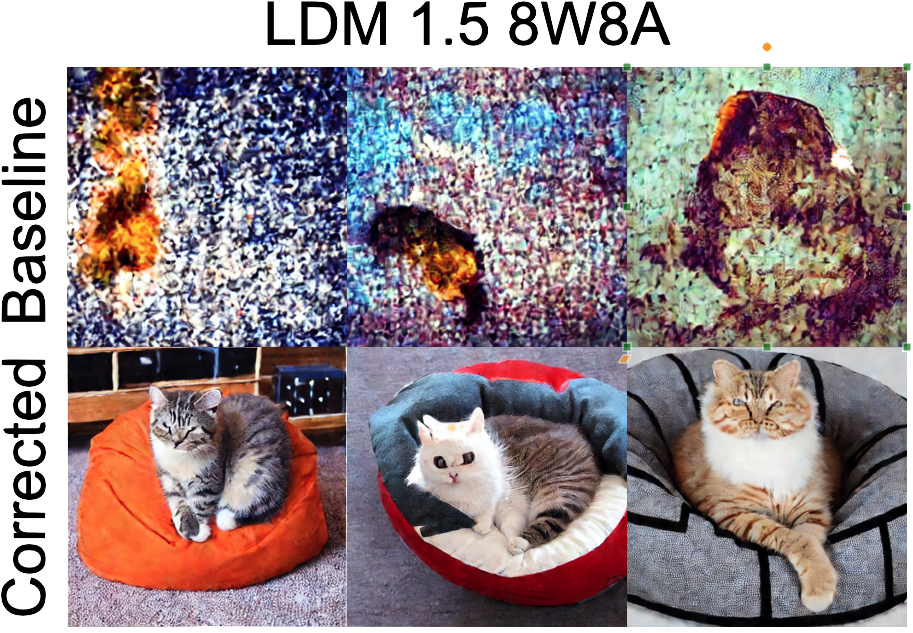}
     \end{subfigure}
     \hfill
     \begin{subfigure}[b]{0.22\textwidth}
         \centering
        \includegraphics[width=\textwidth]{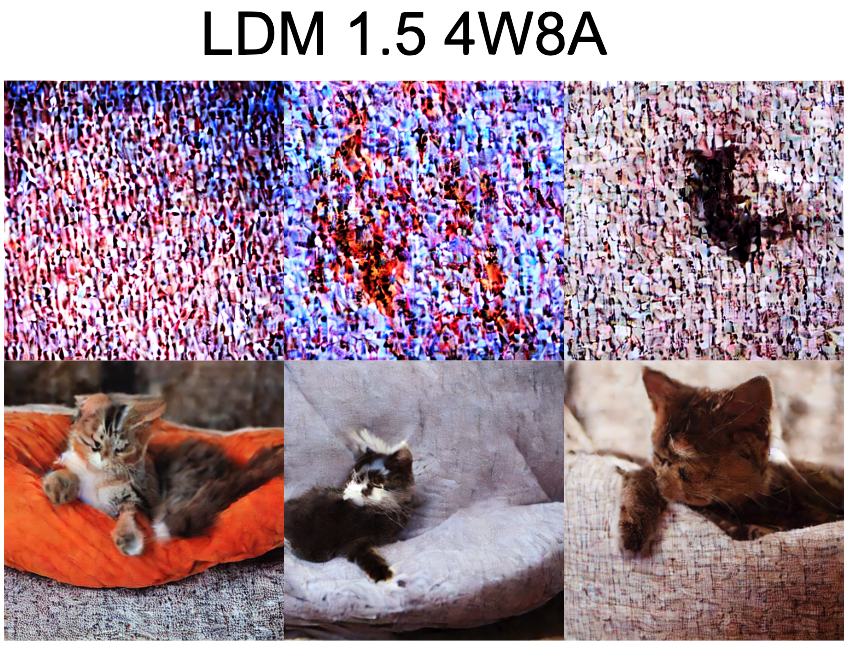}
     \end{subfigure}
     \hfill
     \begin{subfigure}[b]{0.22\textwidth}
         \centering
         \includegraphics[width=\textwidth]{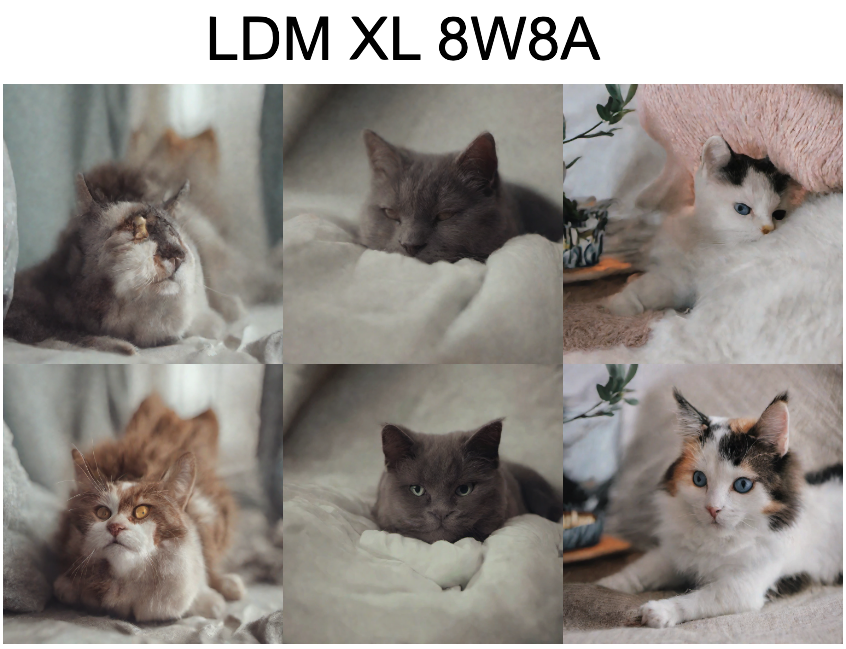}
     \end{subfigure}
     \hfill
     \begin{subfigure}[b]{0.22\textwidth}
         \centering
         \includegraphics[width=\textwidth]{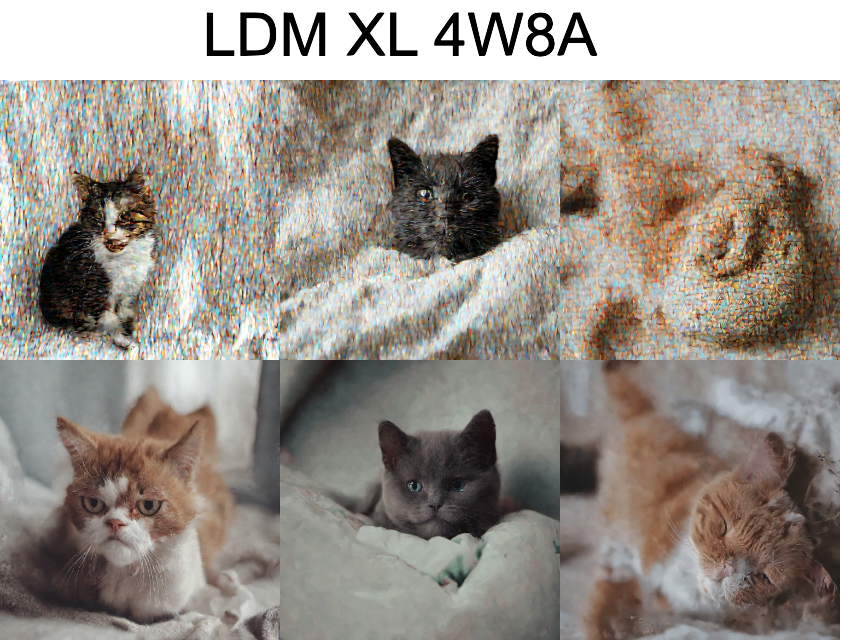}
     \end{subfigure}
        \caption{Local Noise Correct Quantization Strategy.}
        \label{fig:local_examples}
\end{figure*}

\subsection{Qualitative Examples}
We present qualitative examples for both global hybrid strategy Figure \ref{fig:global_examples} and local noise correction strategy Figure \ref{fig:local_examples}. Qualitatively, LDM XL is more robust to min-max quantization but LDM 1.5 generates extensively corrupted images with naive min-max quantization. Nonetheless, the local noise corrections can significantly improve the quality of the images by just identifying 10\% of most sensitive modules.

\section{Ablations}
\label{sec:ablations}
We perform all ablations on LDM 1.5 8W8A quantization setting for consistency.

\begin{table*}[h]
\begin{tabular}{|c|c|cccc|c|cccc|c|}
\hline
      & in\_blocks & \multicolumn{4}{c|}{down\_blocks}                                                                & mid\_block & \multicolumn{4}{c|}{up\_blocks}                                                               &             \\ \hline
      &            & \multicolumn{1}{c|}{0}      & \multicolumn{1}{c|}{1}      & \multicolumn{1}{c|}{2}      & 3      &            & \multicolumn{1}{c|}{0}      & \multicolumn{1}{c|}{1}     & \multicolumn{1}{c|}{2}     & 3     & out\_blocks \\ \hline
TBOPs & 173.63     & \multicolumn{1}{c|}{161.00} & \multicolumn{1}{c|}{148.98} & \multicolumn{1}{c|}{136.91} & 135.46 & 133.14     & \multicolumn{1}{c|}{128.19} & \multicolumn{1}{c|}{99.34} & \multicolumn{1}{c|}{68.97} & 43.64 & 43.63       \\ \hline
FID   & 18.54      & \multicolumn{1}{c|}{18.15}  & \multicolumn{1}{c|}{17.98}  & \multicolumn{1}{c|}{17.54}  & 17.56  &     17.69       & \multicolumn{1}{c|}{16.98}       & \multicolumn{1}{c|}{16.94}      & \multicolumn{1}{c|}{\textbf{16.58}} & 21.31 & 27.97       \\ \hline
\end{tabular}
\caption{Progressively quantize blocks leads to fewer BOPs. The position of the best hybrid configuration matches the block sensitivity.}
\label{tbl:cofig}
\end{table*}

\subsection{Hybrid Quantization Configurations}
As mentioned in Section \ref{subsec:global}, we examine the different hybrid quantization configurations to verify the block sensitivity shown in Figure \ref{fig:block_sensitivity}. We explore all hybrid quantization configurations on LDM 1.5 and evaluate FID and computational efficiency in Table \ref{tbl:cofig}. Though FID suggests that higher-precision quantization generates less presentable images, which contrasts to the expectation. We argue that FID is not an optimal metric to quantize the quantization loss when the quantization noise is low. Only when significant quantization noise corrupts the image, a high FID indicates the generated images are too noisy.

\subsection{Local Quantization Corrections}
As we obtain the local sensitivity for the LDM, we can identify the sensitive local modules from most sensitive to least sensitive based on $\mathbf{SQNR}_{\xi}$. We explore the portion of the modules to adapt SmoothQuant and evaluate $\mathbf{\overline{SQNR}}_{\theta}$. Note that both statistic preparation and computation on $diag(\mathbf{s})^{-1}$ proportionally increase with the number of identified modules. As the 10\% most sensitive modules identified, SmoothQuant can effectively address the activation quantization challenge as $\mathbf{\overline{SQNR}}_{\theta}$ increases and saturate till 90\%. And smoothing least sensitive modules results in deterioration due to unnecessary quantization mitigation. With different mitigation factor, $\alpha$, the behaviour remains consistent, only higher $\alpha$ reduces quantization errors by alleviating activation quantization challenge.

\begin{figure}[htbp]
    \centering
    \includegraphics[scale=0.2]{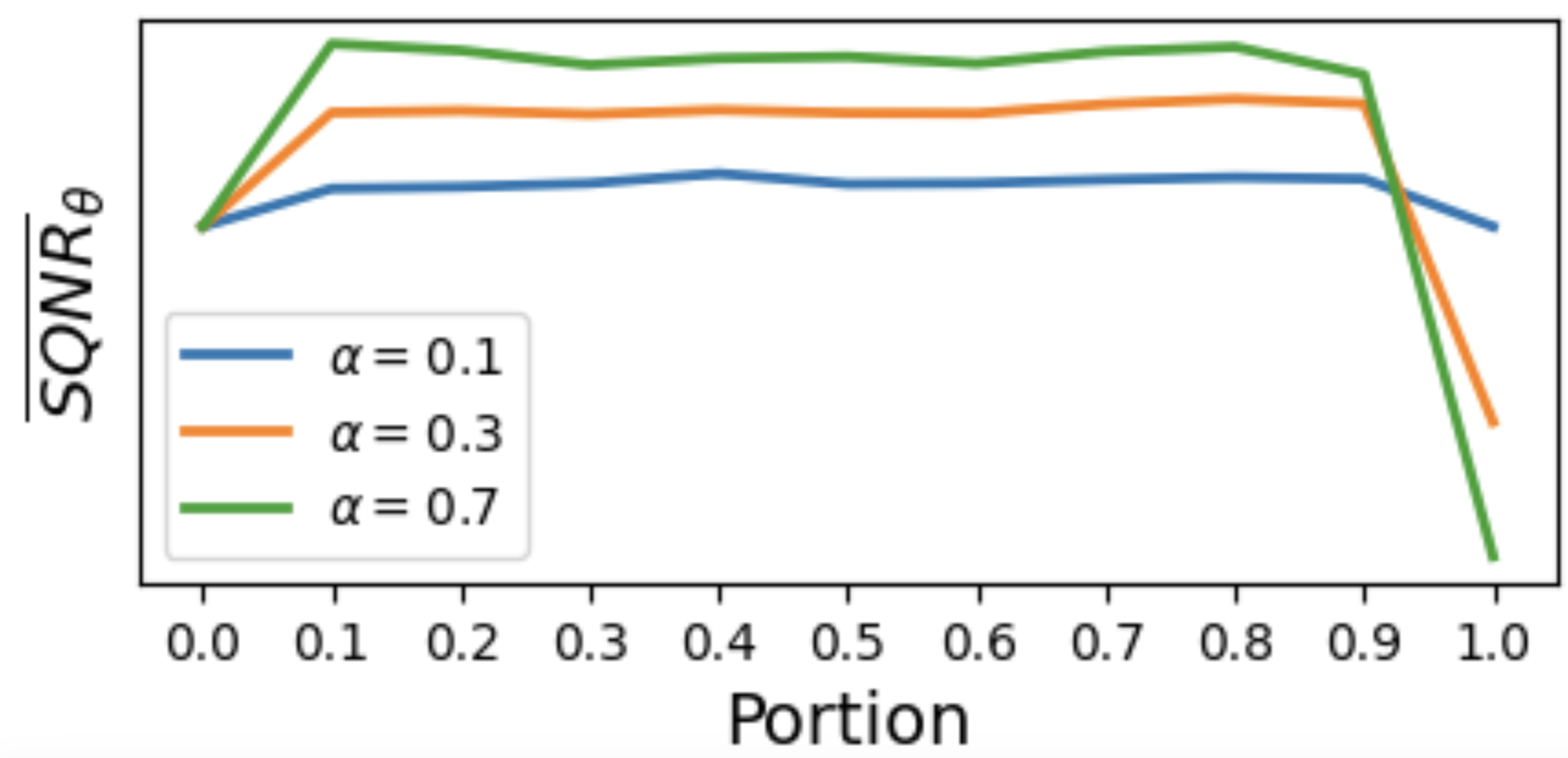}
    \caption{Smoothing 10\% most sensitive modules results in sizable increase in $\mathbf{\overline{SQNR}}_{\theta}$.}
    \label{fig:portion_ablation}
\end{figure}

We also adapt SmoothQuant to the hybrid quantization to evaluate the benefit of such activation quantization noise correction. 10\% most sensitive modules are identified and applied with SmoothQuant ($\alpha=0.7$). FID and $\mathbf{\overline{SQNR}}_{\theta}$ are reported. 1-step calibration significantly reduces the quantization noise for the hybrid quantization. However, SmoothQuant does not make visible difference. This is because identified sensitive modules reside mostly in the upsampling blocks, which are replaced with higher-precision quantization and therefore the quantization noise is already reduced by the hybrid strategy.
\begin{table}[htbp]
\centering
\begin{tabular}{|l|l|l|}
\hline
                     & FID   & $\mathbf{\overline{SQNR}}_{\theta}$  \\ \hline
hybrid               &   27.31    &   17.16    \\ \hline
+ 1-step calibration & 16.58 & 20.45 \\ \hline
++ SmoothQuant       &   16.75    &    20.41   \\ \hline
\end{tabular}
\end{table}

\subsection{Number of Calibration Steps}
As we discuss in \ref{subsub:steps}, we notice that calibrate on multiple sampling steps is problematic for quantization due to gradually changing activation ranges. And quantization parameters are most robust to the added diffusion noise at first sampling step (last forward diffusion step). Hence calibrating on fewer sampling steps leads to less vibrant activation ranges, which results in better quantization calibration. We evaluate the effect on the number of calibration steps on the relative quantization error, $\mathbf{\overline{SQNR}}_{\theta}$, and FID. It is obvious that the proposed 1-step calibration significantly improves the calibration efficiency ($\times{50}$) compared to calibration methods proposed in \cite{li2023q,he2023ptqd,shang2023post}, where calibration data are sampled from all 50 sampling steps.

\begin{table}[htbp]
\centering
\begin{tabular}{|c|c|c|c|}
\hline
   Steps & \begin{tabular}[c]{@{}c@{}}Calibration\\ Efficiency\end{tabular} & FID & $\mathbf{\overline{SQNR}}_{\theta}$ \\ \hline
50 & $\times{1}$                                                      &  213.72   &   11.04   \\ \hline
2 & $\times{25}$                                                    &  47.892   &   13.55   \\ \hline
1  & $\times{50}$                                                     &   27.971  &  14.05    \\ \hline
\end{tabular}
\caption{}
\label{tbl:steps}
\end{table}

\subsection{Efficient metric Computation}
We develop an efficient computation of $\mathbf{SQNR}$ by forward passing two same inputs to all input modules. And we rearrange the output of all modules so that first half of the output is full-precision output, and the second half of the output is quantized output. In this way, $\mathbf{SQNR}$ can be computed at all modules with a single forward pass and with a small number of examples. Other complicated optimization like Q-Diffusion or PTQ4DM collects thousands of calibration data over all 50 sampling steps. Detailed description is included in Supplementary \ref{supp:computesqnr}.

\section{Conclusions}
In summary, this work introduces an innovative quantization strategy, the first of its kind, providing effective quantization solutions for Latent Diffusion Models (LDMs) at both global and local levels by analyzing relative quantization noise, namely $\mathbf{SQNR}$, as the pivoting metric for sensitivity identification. The proposed hybrid quantization solution at global levels identifies a specific LDM block for higher-precision quantization, so that it mitigates relative quantization noise. At local levels, a smoothing mechanism targets the most sensitive modules to alleviate activation quantization noise. Additionally, a single-sampling-step calibration is proposed, leveraging the robustness of local modules to quantization when diffusion noise is strongest in the last step of the forward diffusion process. The collected quantization strategy significantly improves quantization efficiency, marking a notable contribution to LDM optimization strategies.

{
    \small
    \bibliographystyle{ieeenat_fullname}
    \bibliography{main}
}

\clearpage
\appendix
\onecolumn

\section{Module Sensitivity}
\label{supp:modulesensitivity}
We show module sensitivity for LDM 2.1 and LDM XL here. The quantization setting is 8W8A for both models.
\begin{figure}[htbp]
    \centering
    \begin{subfigure}[b]{0.44\textwidth}
    \includegraphics[width=\textwidth]{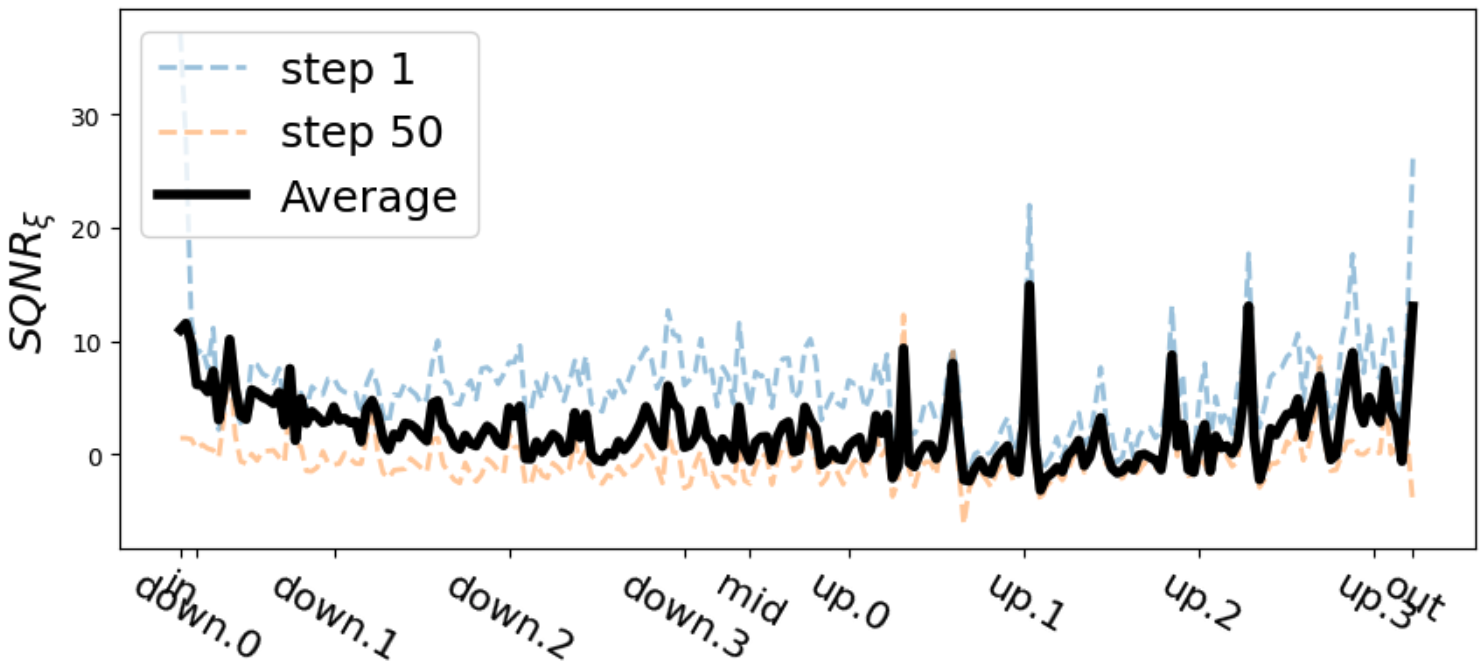}
    \caption{LDM 2.1}
    \label{fig:module_ldm_2.1}
    \end{subfigure}
    \hfill
    \begin{subfigure}[b]{0.44\textwidth}
    \includegraphics[width=\textwidth]{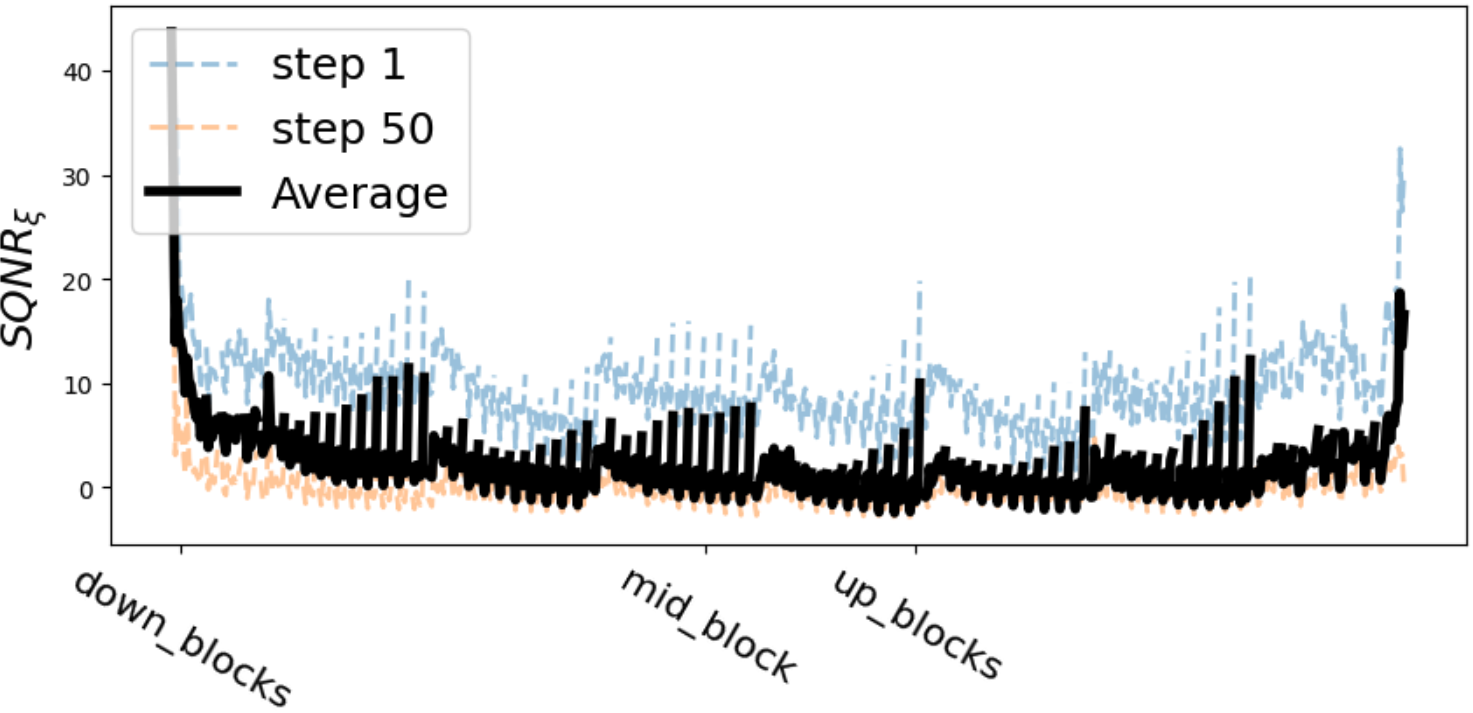}
    \caption{LDM XL}
    \label{fig:module_ldm_xl}
    \end{subfigure}
    \caption{Module sensitivity for LDM 2.1 and LDM XL.}
    \label{fig:module}
\end{figure}

We notice that LDM XL has more robust modules compared to LDM 1.5 and LDM 2.1. We deduce that cross-attention modules are switched to middle section of the model instead of the head-and-tail sections. The cross-attention modules fluactuate the quantization sensitivity because different quantization tolerances between the text-input modules and image-input modules. But the general trend remains similar in all three models that sensitive modules can be identified as they are associated with low $\mathbf{SQNR}$. And the quantization parameters are more robust in the first step of inference.

We also list the 5\% most sensitive modules in all three models.

\begin{lstlisting}
\% LDM 1.5
'down_blocks.1.downsamplers.0.conv',
'up_blocks.1.attentions.2.transformer_blocks.0.ff.net.2',
'down_blocks.2.resnets.0.conv_shortcut',
'down_blocks.1.attentions.1.transformer_blocks.0.ff.net.2',
'up_blocks.1.attentions.1.transformer_blocks.0.ff.net.2',
'mid_block.attentions.0.transformer_blocks.0.ff.net.2',
'up_blocks.1.attentions.0.transformer_blocks.0.ff.net.2',
'up_blocks.2.resnets.0.conv_shortcut', 'up_blocks.1.resnets.2.conv_shortcut',
'up_blocks.1.upsamplers.0.conv', 'mid_block.attentions.0.proj_out',
'down_blocks.2.attentions.0.transformer_blocks.0.ff.net.2',
'up_blocks.1.resnets.1.conv_shortcut', 'down_blocks.2.downsamplers.0.conv'

\% LDM 2.1
'up_blocks.2.attentions.0.transformer_blocks.0.ff.net.2',
'up_blocks.1.attentions.2.proj_out', 'up_blocks.2.attentions.0.proj_out',
'up_blocks.2.resnets.0.conv2',
'up_blocks.1.attentions.1.transformer_blocks.0.ff.net.2',
'up_blocks.1.attentions.2.transformer_blocks.0.ff.net.2',
'up_blocks.2.resnets.1.conv1', 'up_blocks.2.attentions.1.proj_out',
'up_blocks.3.attentions.0.transformer_blocks.0.ff.net.2',
'up_blocks.2.upsamplers.0.conv', 'up_blocks.2.resnets.1.conv_shortcut',
'up_blocks.2.resnets.2.conv1', 'up_blocks.2.resnets.2.conv_shortcut',
'up_blocks.2.attentions.2.proj_out'

\% LDM XL
'down_blocks.2.attentions.1.transformer_blocks.3.attn2.to_out.0',
'down_blocks.2.attentions.1.transformer_blocks.2.attn2.to_out.0',
'down_blocks.2.attentions.1.transformer_blocks.4.attn2.to_out.0',
'down_blocks.2.attentions.1.transformer_blocks.5.attn2.to_out.0',
'down_blocks.2.attentions.0.transformer_blocks.9.attn2.to_out.0',
'down_blocks.2.attentions.1.transformer_blocks.7.attn2.to_out.0',
'down_blocks.2.attentions.1.transformer_blocks.8.attn2.to_out.0',
'down_blocks.2.attentions.1.transformer_blocks.9.attn2.to_out.0',
'down_blocks.2.attentions.1.transformer_blocks.1.attn2.to_out.0',
'up_blocks.0.attentions.0.transformer_blocks.2.attn2.to_out.0',
'up_blocks.0.attentions.1.transformer_blocks.3.attn2.to_out.0',
'down_blocks.2.attentions.0.transformer_blocks.1.attn2.to_out.0',
'down_blocks.2.attentions.1.transformer_blocks.6.attn2.to_out.0',
'up_blocks.0.attentions.1.transformer_blocks.5.attn2.to_out.0',
'up_blocks.0.attentions.1.transformer_blocks.7.attn2.to_out.0',
'down_blocks.2.attentions.1.transformer_blocks.0.attn2.to_out.0',
'up_blocks.0.attentions.1.transformer_blocks.2.attn2.to_out.0',
'up_blocks.0.attentions.0.transformer_blocks.1.attn2.to_out.0',
'up_blocks.0.attentions.0.transformer_blocks.4.attn2.to_out.0'
\end{lstlisting}
For LDM 1.5 and 2.1, three kinds of operations are identified as most sensitive: a). Sampling operations; b). Transformer output projection; c). Shortcut connection. But for LDM XL, the transformer projection consistently deteriorates the quantization noise. There can be a specific solution to LDM XL for this consistent deterioration. But the smoothing mechanism can address the problem to a extend.

\section{Quantization Optimization}
\label{supp:quantizeoptimize}
We list the number of samples, number of inference steps required for different optimized quantization and compare with our proposed method. While other optimized methods require 50 calibration steps, we only require 1 step for activation range register, and 1 step for calibration. Though other complicated quantizations benefit the quality as indicated by higher $\mathbf{SQNR}$, our proposed method is highly efficient and the quality is on-par compared with others.
\begin{table}[htbp]
\centering
\begin{tabular}{|l|l|l|l|l|}
\hline
           & No. Samples & No. Steps & Efficiency    & $\mathbf{SQNR}$ \\ \hline
Entorpy    & 512         & 50        & $\times{2}$   & 18.02           \\ \hline
Percentile & 256         & 50        & $\times{4}$   & 17.13           \\ \hline
BREQ       & 1024        & 50        & $\times{1}$   & 18.86           \\ \hline
Our        & 256         & 1+1       & $\times{100}$ & 17.78           \\ \hline
\end{tabular}
\end{table}

\section{Efficient Computation of SQNR}
\label{supp:computesqnr}
To efficiently compute the relative metric $\mathbf{SQNR}$, we aim to compute the score on the fly without storing any computational expensive feature maps. This means we compute the score with a single forward pass. To realize this efficient computation, the model $\theta_{*}$ should have a switch that controls the quantized and unquantized states. Two identical inputs are passed to every input node and each will only flow through quantized state or unquantized state afterwards. For every module, it is therefore possible to compute the relative quantization noise at the output instantaneously. A pseudo code is shown as following:
\begin{lstlisting}[language=Python]
def SQNR(unquantized, quantized):
    Ps = torch.norm(unquantized.flatten(1), dim=1)
    Pn = torch.norm((unquantized - quantized).flatten(1), dim=1)
    return 20 * torch.log10(Ps / Pn)

def compute_module_SQNR(module,input):
    if module is (text input node or image input node after 1st sampling step):
        chunks = input.chunk(4) # [quantized_cond,unquantized_cond,
                                # quantized_uncond,unquantized_uncond]
        quantized_input = concatenate([chunks[0],chunks[2]])
        unquantized_input = concatenate([chunks[1],chunks[3]])
    else:
        quantized_input,unquantized_input = input.chunk(2)

    quantized_output = module(quantized_input,quantized=True)
    unquantized_output = module(unquantized_input,quantized=False)

    store(module.name,SQNR(unquantized_output,quantized_output).mean())

    output = concatenate([quantized_output,unquantized_output])
    if module is the final output module:
        chunks_ = output.chunk(4)
        output = concatenate([chunks_[0],chunks_[2],chunks_[1],chunks_[3]])
    return output
\end{lstlisting}

\begin{figure}[h]
    \centering
    \includegraphics[scale=0.35]{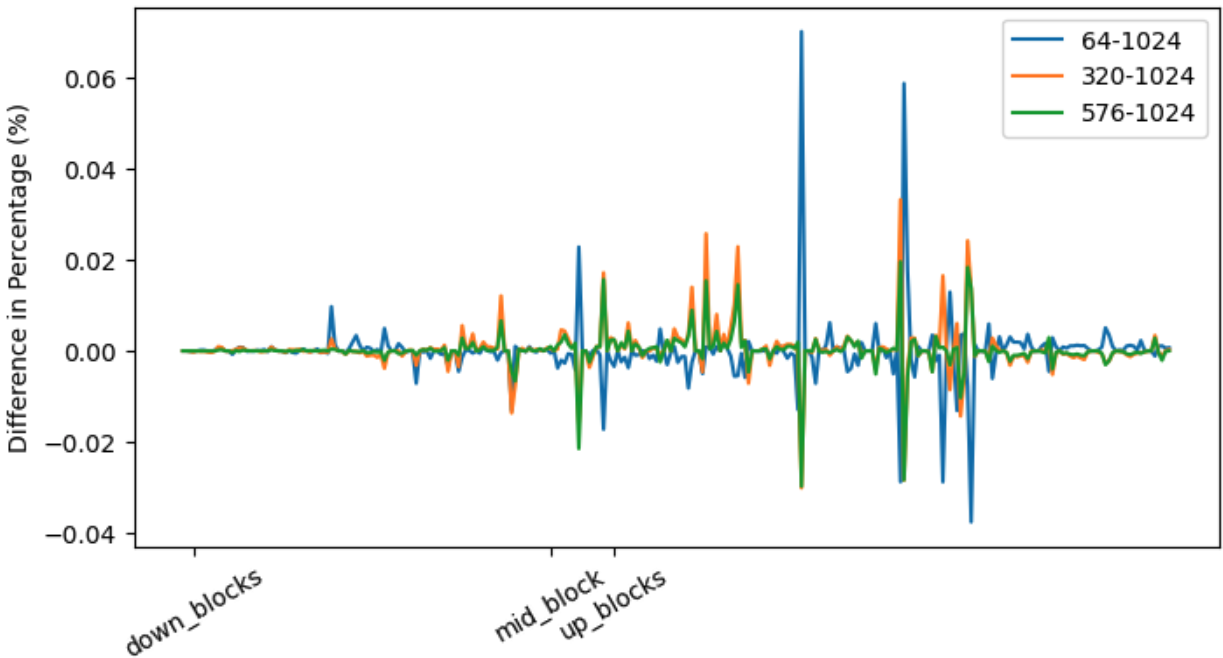}
    \caption{Computing SQNR for all modules with different numbers of samples. The relative differences are small between using 1024 samples and 64 samples.}
    \label{fig:diff_steps}
\end{figure}

Though the duplicated input halves the maximum number of executable samples per batch, we argue the efficiency is more beneficial for our proposed strategy and we justify that we only need a small number of examples to compute a compelling $\mathbf{SQNR}$.We measure the time-averaged relative difference between $\mathbf{SQNR}$ computed with 1024 samples and that computed with 64 samples for all modules. The result is shown in Figre \ref{fig:diff_steps}. The small relative difference (~0.07\% at maximum) suggests that computing $\mathbf{SQNR}$ with just 64 samples is comparable with 1024 samples.

\section{LDM Configurations}
\label{supp:ldm_config}
\begin{table}[htbp]
\centering
\begin{tabular}{|l|l|l|l|}
\hline
        & Image Size & Inference Steps & Sampler                \\ \hline
LDM 1.5 & 512, 512   & 50              & PNDMScheduler          \\ \hline
LDM 2.1 & 512, 512   & 50              & PNDMScheduler          \\ \hline
LDM XL  & 1024, 1024 & 50              & EulerDiscreteScheduler \\ \hline
\end{tabular}
\end{table}

\section{Activation Outliers}
\label{supp:outliers}
We visualize activation ranges in 3d for 4 examples at three operations we mentioned. The red region suggests high activation range and blue region suggests low activation range. As mentioned, the outlier in certain channels make the min-max quantization difficult.

\begin{figure}
    \centering
    \begin{subfigure}[b]{0.9\textwidth}
    \includegraphics[width=\textwidth]{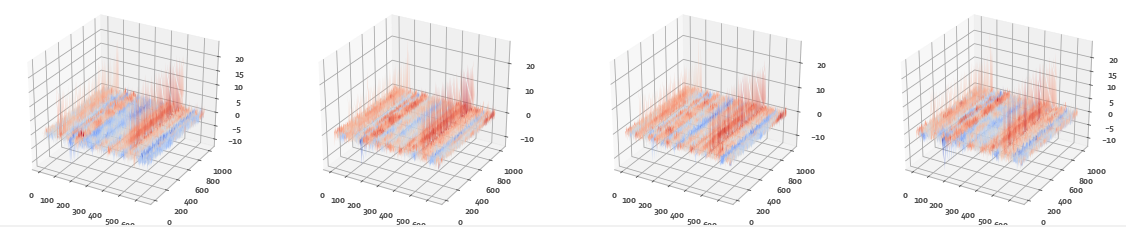}
    \caption{3D Activation Range for Sampling Operation}
    \label{fig:sampling3d}
    \end{subfigure}
    \hfill
    \begin{subfigure}[b]{0.9\textwidth}
    \includegraphics[width=\textwidth]{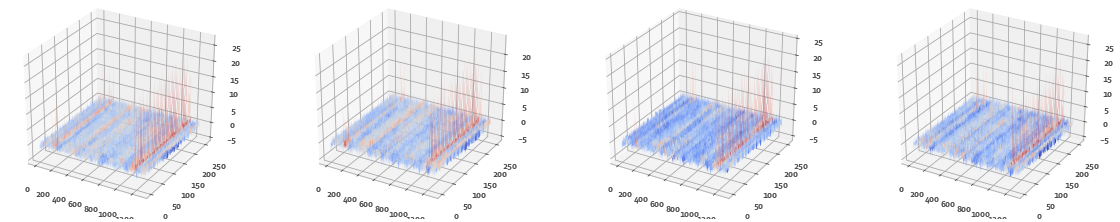}
    \caption{3D Activation Range for Transformer Output Projection}
    \label{fig:project3d}
    \end{subfigure}
    \hfill
    \begin{subfigure}[b]{0.9\textwidth}
    \includegraphics[width=\textwidth]{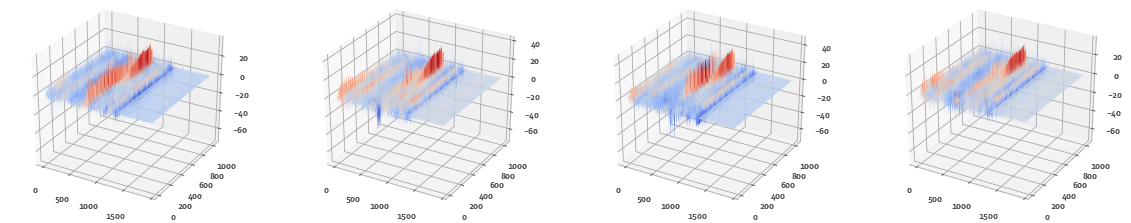}
    \caption{3D Activation Range for Shortcut Connection Operation}
    \label{fig:shortcut3d}
    \end{subfigure}
    \hfill
    \caption{4 examples for three operations.}
    \label{fig:3dvisual}
\end{figure}

\end{document}